\documentclass[11pt]{article}

\usepackage{microtype} 
\usepackage{booktabs}  
\usepackage{url}  
\usepackage{csquotes}

\usepackage{amsmath}
\usepackage{amsthm}

\usepackage[final]{automl}

\usepackage{booktabs}       
\usepackage{nicefrac}       
\usepackage{microtype}      
\usepackage{xcolor}         

\usepackage{natbib}
\usepackage{color}
\usepackage{amsmath}
\usepackage{amsthm}
\usepackage{bm}

\usepackage{graphicx}
\usepackage{wrapfig}

\usepackage{multirow}
\usepackage{enumitem}

\usepackage{algorithm}
\usepackage{algorithmic}

\newtheorem{prop}{\bf Proposition}

\newcounter{myfootnotecounter}

\theoremstyle{plain}

\theoremstyle{definition}

\theoremstyle{remark}

\usepackage[textsize=tiny]{todonotes}

\usepackage{natbib}
\usepackage{filecontents}

\title{Poisson Process for Bayesian Optimization}

\author[1$\dagger$]{\nameemail{Xiaoxing Wang}
{figure1\_wxx@sjtu.edu.cn}}
\author[2$\dagger$]{\nameemail{Jiaxing Li}{810992213@qq.com}}
\author[3$*$]{\nameemail{Chao Xue}{xuechao19@jd.com}}
\author[4]{\nameemail{Wei Liu}{lwwd1996@163.com}}
\author[2]{\nameemail{Weifeng Liu}{liuwf@upc.edu.cn}}
\author[1]{\nameemail{Xiaokang Yang}{xkyang@sjtu.edu.cn}}
\author[1]{\nameemail{Junchi Yan}{yanjunchi@sjtu.edu.cn}}
\author[4]{\nameemail{Dacheng Tao}{dacheng.tao@gmail.com}}

\affil[1]{Department of Computer Science and Engineering, and MoE Key Lab of Artificial Intelligence, Shanghai Jiao Tong University}
\affil[2]{College of Control Science and Engineering, China University of Petroleum (East China)}
\affil[3]{JD Explore Academy}
\affil[4]{The University of Sydney}
\affil[$\dagger$]{Equal Contributions. This work was done while the authors were interns at JD Explore Academy.}
\affil[$^*$]{Corresponding Author}

\hypersetup{%
  pdfauthor={AutoML}, 
  pdftitle={Documentation for the automl Package},
  pdfsubject={Documentation for automl Package},
  pdfkeywords={AutoML, LaTeX, style}
}

\begin{document}

\maketitle

\begin{abstract}
Bayesian Optimization (BO) is a sample-efficient black-box optimizer, and extensive methods have been proposed to build the absolute function response of the black-box function through a probabilistic surrogate model, including Tree-structured Parzen Estimator (TPE), random forest (SMAC), and Gaussian process (GP). However, few methods have been explored to estimate the relative rankings of candidates, which can be more robust to noise and have better practicality than absolute function responses, especially when the function responses are intractable but preferences can be acquired.
To this end, we propose a novel ranking-based surrogate model based on the Poisson process and introduce an efficient BO framework, namely {Poisson Process Bayesian Optimization} (PoPBO). Two tailored acquisition functions are further derived from classic LCB and EI to accommodate it.
Compared to the classic GP-BO method, our PoPBO has lower computation costs and better robustness to noise, which is verified by abundant experiments. The results on both simulated and real-world benchmarks, including hyperparameter optimization (HPO) and neural architecture search (NAS), show the effectiveness of PoPBO.
\end{abstract}

\section{Introduction} \label{sec:intro}
Bayesian  optimization (BO)~\citep{mockus1978application} is a popular black-box optimization paradigm and has achieved great success in a number of challenging fields, such as robotic control~\citep{DBLP:robotic_journals/amai/CalandraSPD16}, biology~\citep{bo_bio_genedesign}, and hyperparameter tuning for complex learning tasks~\citep{DBLP:tpe_conf/nips/BergstraBBK11}.
A standard BO routine usually consists of two alternate steps: 1) Train a probabilistic surrogate model to build the response surface of a black-box function $f(x)$; 2) Suggest the next query by optimizing an acquisition function according to the learned response surface. 
Popular surrogate models for the first step include random forest (SMAC)~\citep{DBLP:smac_conf/lion/HutterHL11}, Tree-structure Parzen Estimator (TPE)~\citep{DBLP:tpe_conf/nips/BergstraBBK11,bore,likelihoodfree}, Gaussian Process (GP)~\citep{DBLP:spearmint_conf/nips/SnoekLA12} and Bayesian Neural Network (BNN)~\citep{DBLP:BOHAMIANN_conf/nips/SpringenbergKFH16, DBLP:dngo_conf/icml/SnoekRSKSSPPA15}, aiming to estimate the distribution of function values for each candidate. 
Classic acquisition functions for the second step include Expected Improvement (EI)~\citep{DBLP:ei_journals/jgo/Mockus94}, Thompson Sampling (TS)~\citep{ DBLP:thompson_conf/nips/ChapelleL11, DBLP:thompson_conf/icml/AgrawalG13} and Upper/Lower Confidence Bound (UCB/LCB)~\citep{DBLP:ucb_journals/tit/SrinivasKKS12}, aiming at the exploration and exploitation trade-off. 
Most of the prior BO methods~\citep{DBLP:tpe_conf/nips/BergstraBBK11,DBLP:smac_conf/lion/HutterHL11,DBLP:spearmint_conf/nips/SnoekLA12} attempt to build an absolute response surface\footnote{In this work, `absolute response' of a black-box function estimate as the raw values at candidate points. `relative 
response' of a black-box function estimates the ranking of candidate points, which can be computed by pairwise comparing.} of a black-box function based on the observed function values, and few have been explored to build a relative response surface.

However, absolute metrics can have the following shortcomings. \textbf{1) Absolute response can be difficult to obtain or even unavailable in some practical scenarios}, such as sports games and recommender systems where only relative response can be provided by pairwise comparison~\cite{gnnrank}. \textbf{2) Absolute response can be sensitive to noise}, as pointed out by~\cite{RankingICDM05}. Such an issue will affect the performance of BO in noisy real-world scenarios. 
\textbf{3) It can be challenging to directly transfer the surrogate models for absolute response surfaces.}
In particular, multi-fidelity metrics usually require multiple absolute responses for the same candidate. It is hard to utilize history observations on a coarse-fidelity metric to warm up the training of surrogate models on a fine-grained-fidelity one. Similarly, in hyperparameter optimization (HPO) and neural architecture search (NAS) tasks, it is hard to transfer the performance across different datasets.

On the contrary, relative metrics can be effective cures for the above issues. \textbf{1) Relative response such as ranking has better practicality} since the information about candidate preferences can be more easily acquired than raw value~\citep{DBLP:PBO_conf/icml/GonzalezDDL17,noabsolute_kahneman2013prospect, DBLP:noabsolute_conf/adaptive/2007, DBLP:PBO_conf/icml/GonzalezDDL17}.
\textbf{2) Relative response is more robust to noise than absolute response} since relation such as ranking between candidates is hard to be disrupted by noise, which is verified by~\cite{DBLP:topk_conf/aaai/NguyenTLJ21, DBLP:shen_conf/icml/SalinasSP20}. 
In this work, we also analyze the better robustness of rankings against absolute response in Sec.~\ref{subsec:ranking_based} under the common additive Gaussian noise assumption. 
\textbf{3) Relative response has better transferability}, such as rankings between candidates, since they are usually comparable among multi-fidelity metrics or evaluations across different datasets for the same candidate. It is also demonstrated by~\citep{DBLP:shen_conf/icml/SalinasSP20,DBLP:topk_conf/aaai/NguyenTLJ21, transfer_feurer2018scalable}.

Some BO methods utilize the relative responses.
Preferential BO~\citep{DBLP:PBO_conf/icml/GonzalezDDL17,DBLP:ppbo_conf/icml/MikkolaTJRK20,bope} attempts to capture the relative preference by pairwise comparison. However, they either rely on a computationally-expensive soft-Copeland score (PBO) or need to optimize EI by the projective preferential query (PPBO) to propose the next query. 
\cite{DBLP:topk_conf/aaai/NguyenTLJ21} extend the pairwise comparison to comparison among $k$ samples, but it requires building an absolute response through the Gaussian Process and capturing the local ranking among a fixed number of neighbors via a multi-nominal logit model.

\begin{figure*}
    \centering
  \begin{subfigure}[t]{0.48\linewidth}
    \centering
    \includegraphics[width=0.98\linewidth]{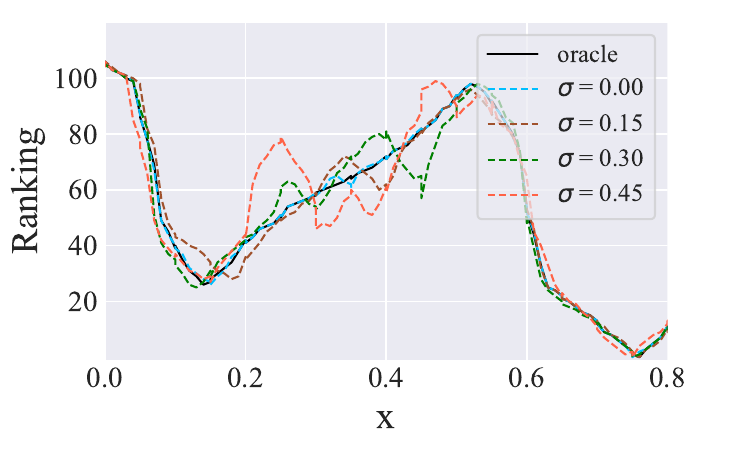}  
    \vspace{-8pt}
    \caption{Response Surface of GP (value-based model)}
    \label{fig:compare_gp}
  \end{subfigure}
  \begin{subfigure}[t]{0.48\linewidth}
    \centering
    \includegraphics[width=0.98\linewidth]{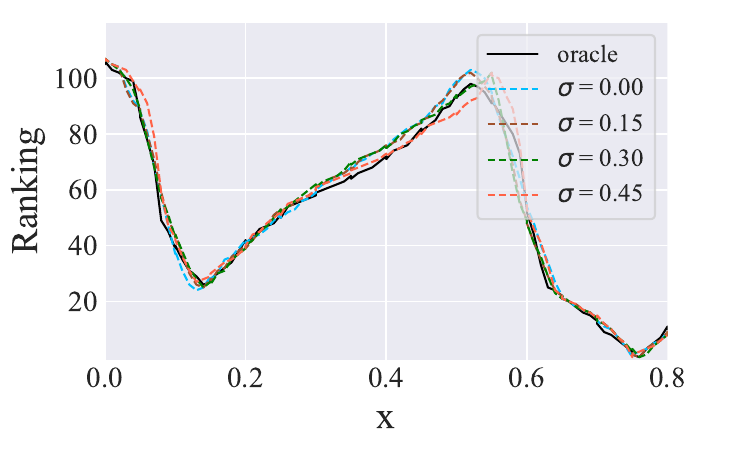} 
    \vspace{-8pt}
    \caption{Response Surface of PP (ranking-based model)}
    \label{fig:compare_pp}
  \end{subfigure}
\vspace{-8pt}
\caption{
We compare the sensitivity of additive Gaussian noise between GP (value-based) response surface and PoPBO (ranking-based response surface) on the Forrester function. Based on the function value, the solid black line (oracle) indicates actual rankings over 100 points evenly spaced from 0 to 0.8. We draw lines between the 100 predictions by linear interpolation for a clear illustration. 
The dashed lines indicate predicted rankings over the 100 points by (a) Gaussian process and (b) Poisson process on observations with varying degrees of noise whose standard deviation $\sigma$ ranges from $0$ to $0.45$. Each response surface is trained on the same 15 queries. Note that GP performs worse as the standard deviation of noise increases. In contrast, PP performs consistently well due to its great robustness against noise.
}
\vspace{-5pt}
\label{fig:compare}
\end{figure*}

In contrast, we propose to capture the global ranking of each candidate among the whole feasible domain (search space) and model the relative response. On the one hand, we can directly search the optimum based on our relative response surface and obtain the next query without computationally expensive procedure. 
On the other hand, we directly build a ranking-based relative response surface without building an absolute response surface first. 
Specifically, we adopt Poisson Process (PP) to capture the global ranking, which is naturally suitable since the ranking of a candidate can be figured out by counting the number of better candidates. 
Verified experiments on the Forrester function with various degrees of additive Gaussian noise are conducted to show the robustness of our response surface capturing the global ranking. Comparison with GP-BO is illustrated in Fig.~\ref{fig:compare}. The detailed settings can be found in the caption and Appendix~\ref{supp:sub:forrester}.
Our contributions can be summarized as follows:

\textbf{1) Ranking-based Response Surface based on Poisson Process.} 
Unlike the prior absolute response surface~\citep{DBLP:tpe_conf/nips/BergstraBBK11,DBLP:spearmint_conf/nips/SnoekLA12}, nor those~\citep{DBLP:topk_conf/aaai/NguyenTLJ21} using relative evidence likelihood based on absolute responses,
this work is the first to directly capture the global ranking over a feasible domain via Poisson process. The robustness against noise is also analyzed in Sec.~\ref{subsec:ranking_based} and illustrated in Fig.~\ref{fig:compare}. 

\textbf{2) Tailored Acquisition Function for Ranking-based Response Surface.}
Two acquisition functions for our response surface, named R-LCB and ERI, are deduced from the vanilla LCB and EI for better exploitation-exploration trade-off. Gradients of the proposed acquisition functions w.r.t. candidates are also derived, so the next query can be optimized by ADAM.

\textbf{3) Computationally-efficient Bayesian Optimization Framework.}
The proposed ranking-based response surface and acquisition functions form an efficient and novel Bayesian optimization framework: {Poisson Process Bayesian Optimization} (PoPBO). The computational complexity of PoPBO is $O(N^2)$, much lower than that of GP-BO ($O(N^3)$). Comparison is shown in Fig.~\ref{fig:time_cost}. 

\textbf{4) Extensive Empirical Study with Strong Performance.} Our method outperforms many prior BO methods on both simulated functions and real-world benchmarks, including HPO and NAS.

\section{Preliminaries and Background} \label{sec:related_work}
\textbf{Bayesian Optimization.}
Consider minimizing a black-box function $x^*=\mathop{\arg\min}_{x\in{X}} {f(x)}$, where $f(\cdot)$: $X \to \mathbb{R}$, defined on a d-dimensional feasible domain $X \subset \mathbb{R}^d$. The observations have additive noise. 
Bayesian optimization (BO) is an efficient method to solve such a problem, especially when the black-box function is expensive to be evaluated and has no closed-form expression~\citep{DBLP:no_closed_func, DBLP:ablr}. It alternately trains a surrogate model to estimate the response surface of the black-box function based on the observed samples and suggest the next query based on a acquisition function balancing the exploitation and exploration. 
Gaussian process is one of the most classic and popular surrogate models, which assumes a GP prior to the black-box function and computes the posterior conditioned on the observations. 
SMAC~\citep{DBLP:smac_conf/lion/HutterHL11} introduces random forests for regression, which can be used to handle categorical hyperparameters. 
TPE~\citep{DBLP:tpe_conf/nips/BergstraBBK11} models two densities for each sample: $l(x) = p(y < \alpha|x, D)$ and $g(x) = p(y > \alpha|x, D)$ via kernel density estimator, and then optimize the ratio $l(x) / g(x)$ to suggest the next query. 
Recently, Bayesian neural network is also introduced into BO framework~\citep{DBLP:dngo_conf/icml/SnoekRSKSSPPA15} to estimate the response surface. \citet{DBLP:BOHAMIANN_conf/nips/SpringenbergKFH16} improves the robustness by evaluating the posterior via a stochastic gradient MCMC method~\citep{DBLP:sghmc_conf/icml/ChenFG14}. 

After building the response surface via a surrogate model, BO suggests the next query via an acquisition function considering the trade-off between exploitation and exploration. 
Popular acquisition functions include expected improvement, Thompson sampling, and upper/lower confidence bound due to their ease of use and strong performance.

\textbf{BO with Relative Metrics.}
The relative metric does not have to utilize absolute responses of the black-box function.
Some methods focus on the cases where the function evaluation is not directly accessible~\citep{DBLP:noabsolute_conf/adaptive/2007, DBLP:PBO_conf/icml/GonzalezDDL17,DBLP:ppbo_conf/icml/MikkolaTJRK20,DBLP:PBBO_conf/mlsp/SiivolaDAGMV21}. 
Absolute responses can be difficult to obtain or even unavailable in some practical scenarios, such as sports games and recommender systems~\citep{DBLP:noabsolute_conf/adaptive/2007} where only relative evaluation can be provided by pairwise comparisons.
\textit{Preferential Bayesian Optimization} (PBO)~\citep{DBLP:PBO_conf/icml/GonzalezDDL17} captures correlations between different inputs to find the optimal value of a latent function, which requires limited comparisons. To handle a high-dimensional black-box function, \textit{Projective Preferential Bayesian Optimization} (PPBO)~\citep{DBLP:ppbo_conf/icml/MikkolaTJRK20} proposes a projective preferential query allowing for the feedback given by human interaction. However, they ignore the tie situations and have to rely on a computationally expensive procedure to suggest the next query. \citet{DBLP:topk_conf/aaai/NguyenTLJ21} extend the above method by comparing k samples but have to model the absolute response surface by Gaussian Process and assume the noise obeys Gumbel distribution.
In addition, ranking-based methods~\citep{transfer_feurer2018scalable, DBLP:shen_conf/icml/SalinasSP20} can also facilitate the identification of similar runs for transfer learning, reusing insights from past similar experiments. 
This work, on the contrary, makes the first attempt to directly capture the global ranking of candidates based on Poisson process and derive a novel Bayesian Optimization framework named PoPBO. We analyze the robustness of relative metric (ranking) against noise and show the outstanding performance of our method on various simulated benchmarks and real-world datasets.


\section{Poisson Process for Bayesian Optimization}
\label{sec:ppbo}

\subsection{Ranking-based Metric}
\label{subsec:ranking_based}
Consider a black-box objective function $f(\cdot)$ defined on a feasible domain $X$. 
Given a sample $x\in X$ and a subset of the feasible domain $S\subset X$, we define a set $S_x = \{y| y\in S, f(y)< f(x)\}$, consisting of the better candidates than $x$ in $S$. Hence, we can estimate the superiority of $x$ for $f(\cdot)$ against the points in set $S$ by measuring $S_x$.
Specifically, consider two points $x_1, x_2$, if $S_{x_1}$ has a larger measure value than $S_{x_2}$, it represents that there are more points in $S$ better than $x_1$ compared to those better than $x_2$, so $x_1$ is worse than $x_2$.
We would like to capture the ranking of each candidate $x$ over a domain $S$. However, $S$ usually contains a large number of candidates making it intractable to obtain the true ranking. Therefore, we sample a set $\hat{S}$ from the search space $X$ and attempt to capture the ranking of $x$ over $S\cap\hat{S}$, i.e., $|S_x\cap \hat{S}|$.

\textbf{Robustness Analysis.}
Consider two queries $x_1, x_2$ with observations $y_1, y_2$. Suppose observations of the black-box function are subject to additive Gaussian noisy $y = f(x) + \epsilon, \epsilon\sim\mathcal{N}(0, \sigma^2)$ We assume $f(x_1) < f(x_2)$ without loss of generality, the probability of correctly ranking $x_1, x_2$ is:
\begin{equation}
    P(y_1 < y_2) = P(\epsilon_1-\epsilon_2 < f(x_2)-f(x_1)).
\end{equation}
Since $\epsilon_1, \epsilon_2 \sim\mathcal{N}(0, \sigma^2)$ are independent, $\Delta\epsilon=\epsilon_1-\epsilon_2\sim\mathcal{N}(0, 2\sigma^2)$. According to three-sigma rule of thumb, if $f(x_2)-f(x_1) > \sqrt{2}\sigma$, the probability of correctly ranking $x_1, x_2$ is larger than 82.63\%; If $f(x_2)-f(x_1) > 2\sqrt{2}\sigma$, the probability of correctly ranking $x_1, x_2$ is larger than 97.72\%.
Even if observations are noisy, the ranking of candidates is hard to be disrupted. 

Hence, we conclude the ranking metric is robust to noise and assume the observed ranking noiseless in this work. Results of our ranking-based surrogate model on Forrester function with varying degrees of additive noises confirm our analysis, as shown in Fig.~\ref{fig:compare_pp}. Specifically, despite that no prior of noise is considered on the observed ranking when training the PP (ranking-based) surrogate model, it can still capture the oracle ranking (the line of ground truth in black) properly even with noisy observations. 
In contrast, although the GP (value-based) surrogate model considers the prior of noise, it performs worse.

\begin{algorithm}[h!]
\renewcommand{\algorithmicrequire}{\textbf{Require:}}
\caption{PoPBO: Bayesian Optimization with Poisson Process} \label{alg:popbo}
\begin{algorithmic}[1]
\REQUIRE
\textbf{1)} A function $Rank(\{\cdot\})^\dagger$ to rank samples based on a black-box function $f(\cdot)$;
\textbf{2)} A feasible domain (search space) $X$;
\textbf{3)} An acquisition function $\alpha$; 
\textbf{4)} The number of initial points $N$;
\textbf{5)} the number of total training iterations $T$.\\
\STATE Randomly sample $N$ initial points $\hat{S} := \{x_j\}_{j=1}^N$ from $X$;\\ 
\STATE Initialize the parameters $\theta$ of $\lambda_\xi(x)$;\\
\FOR{$t := from~1~to~T$}
    \STATE Get their rankings $\hat{R} := Rank(\hat{S})$;\\
    \STATE Train $\theta$ based on points $\hat{S}$ and $\hat{R}$ by minimizing Eq.~\ref{eq:neg_log_likelihood} through ADAM;\\
    \STATE Get the next query $x^*$ by minimizing acquisition function $\alpha$;\\
    \STATE Update the set of points $\hat{S}:=\hat{S}\cup\{x^*\}$
\ENDFOR
\STATE Get the best query $x_{opt}$ in history based on $Rank(\hat{S})$;\\
\hspace*{-0.23in}\textbf{Output:} The best point $x_{opt}$ in history. \\
\hspace*{-0.23in}{\small $\dagger$: A sorting function can serve as the $Rank({\cdot})$.}
\end{algorithmic}
\end{algorithm}

\subsection{Capturing the Ranking via Poisson Process}
Given a sample $x$ and a set $\hat{S}$, we utilize a random process $\hat{R}_x(S), \forall S\subset X$ to capture the ranking $x$. Note that $\hat{R}_x(S)$ also depends on $\hat{S}$, we omit it for conciseness. In particular, we define $\hat{R}_x \triangleq\hat{R}_x(X)$ to denote the ranking of $x$ over the whole feasible domain $X$, which is a random variable.
We assume the rankings of $x$ over two disjoint areas are independent, i.e., $\hat{R}_x(S_1) \perp \!\!\! \perp
 \hat{R}_x(S_2), \forall S_1,S_2\subset X, S_1\cap S_2 = \emptyset$ since the function $f(x)$ is black-box. Hence, we can model $\hat{R}_x(S), \forall S\subset X$ as an independent increment counting process. Moreover, $\hat{R}_x(S)$ has the following properties: 1) $\hat{R}_x(\emptyset) = 0$ and 2) $\lim_{\Delta s\rightarrow0}\text{P}(\hat{R}_x(S+\Delta s) - \hat{R}_x(S)\geq2) = 0, \forall S\subset X$. Detailed discussion is provided in Appendix~\ref{supp:sec:assumption_discussion}.
Since the supremum of $\hat{R}_x(S)$ is $|\hat{S}|$, $\hat{R}_x(S)$ obeys a truncated non-homogeneous Poisson process~\citep{yigiter2006right} as Eq.~\ref{eq:non_homo_poisson} with parameter $\lambda(s, x), s\in X$.
\begin{align} 
\label{eq:non_homo_poisson}
\hat{R}_x(S) \sim Poisson\left(\int_S \lambda(s,x)\text{d}s\right).
\end{align}
Hence, the ranking of $x$ over the whole feasible domain is $\hat{R}_x = \hat{R}_x(X)$, the probability of $\hat{R}_x = k$ is:
\begin{align}
     \text{P}\left(\hat{R}_x =k|x, \hat{S}\right)
     &= \frac{\left(\int_X\lambda(s, x)\text{d}s\right)^k}{k!\cdot Z(x)} \exp{\left(-\int_X\lambda(s, x)\text{d}s\right)} 
     = \frac{\left(\lambda_\xi(x)|X|\right)^k}{k!\cdot Z(x)} \exp{\left(-\lambda_\xi(x)|X|\right)}, \label{eq:norm_prob} 
\end{align}
where $Z(x)=\sum_{k=0}^{|\hat{S}\backslash\{x\}|}\left[\frac{\left(\lambda(\xi, x)|X|\right)^k}{k!} \exp{\left(-\lambda(\xi, x)|X|\right)}\right]$ is the normalized coefficient and $|\hat{S}\backslash\{x\}|$ is the number of samples without $x$. There exists $\xi\in X$ satisfying $\int_X\lambda(s, x)\text{d}s = \lambda_\xi(x)|X|$ according to the mean value theorem for integrals. We can approximate $\lambda_\xi(x)$ by a multi-layer perceptron (MLP) $\lambda_\xi(x; \theta)$ with parameter $\theta$. 

To train $\theta$, we resort to the maximized loglikelihood estimation (MLE). Give $N(\geq2)$ samples $\hat{S}=\{x_j\}_{j=1}^N$, the ranking of each sample over $\hat{S}$ is $\hat{K}=\{\hat{k}_{x_j}\}_{j=1}^N$. Similar to \citep{DBLP:shen_conf/icml/SalinasSP20}, the log-likelihood $\log{L(\hat{K}|\hat{S}; \theta)}$ can be approximated as follows and $\theta$ can be optimized through ADAM.
\begin{align}
\label{eq:neg_log_likelihood}
\sum_{j=1}^N\Bigg\{ \hat{k}_{x_j} \log{\left(\lambda_\xi(x_j; \theta)|X| \right)} - \log{(\hat{k}_{x_j}!)} 
    - \log{\bigg[ \sum_{i=0}^{N-1}\frac{\left(\lambda_\xi(x_j; \theta)|X| \right)^i}{i!} \bigg]} \Bigg\}.
\end{align}

Once $\theta$ is determined after training on the observations $(\hat{S}, \hat{K})$, the ranking of a new sample $x^*$ over the whole feasible domain $X$ can be predicted, where $Z$ is the normalized coefficient by Eq.~\ref{eq:norm_prob}.
\begin{align}
\label{eq:posterior}
    \text{P}\left(\hat{R}_{x^*}(X)=k|\theta, x^*,\hat{S}\right) = &\frac{\big(\lambda_\xi(x^*; \theta)|X|\big)^k}{k!\cdot Z(x^*)} 
    \cdot\exp{\big(-\lambda_\xi(x^*; \theta)|X|\big)}.
\end{align}

The proposed Bayesian optimization framework with Poisson process (PoPBO) is outlined in Alg.~\ref{alg:popbo}. The acquisition function is introduced in the next section.

\section{Tailored Acquisition Functions for PoPBO}
\label{sec:acq_func}

The existing acquisition functions are designed for absolute response surface considering independent mean and variance, which can be improper for our ranking-based response surface since the mean of Poisson distribution is the same as the variance. 
Directly applying these acquisition functions to our PoPBO will cause the over-exploitation issue. To this end, we introduce a series of acquisition functions, named rectified upper confidence bound (R-LCB) and expected ranking improvement (ERI), derived from vanilla LCB and EI, respectively.

\subsection{Rectified Lower Confidence Bound (R-LCB)}
Ranking of the point $x$ obeys Poisson distribution as Eq.~\ref{eq:posterior}, with expectation $\mu(x) = \lambda_\xi(x)|X|\cdot  \frac{\sum_{i=0}^{N-1}\left(\lambda_{\xi}(x)|X|\right)^i/i!}{\sum_{i=0}^{N}\left(\lambda_{\xi}(x)|X|\right)^i/i!}$ and standard deviation $\sigma(x) = \sqrt{\mu(x)}$ according to the property of Poisson distribution. Thus the vanilla LCB of each point is:
\begin{equation}
\label{eq:LCB}
\alpha_{\text{LCB}}(x) = \mu(x) - \beta\sigma(x) =  \sqrt{\mu(x)}\left( \sqrt{\mu(x)}-\beta\right).
\end{equation}
However, Poisson distribution with a large expectation also has a large variance, indicating less confidence in the ranking prediction. The vanilla LCB will be trapped into an over-exploitation issue. Hence, we propose a rectified LCB (R-LCB) to restrict the lower value to a threshold:
\begin{equation}
\label{eq:rLCB}
\alpha_{\text{R-LCB}}(x) = 
\left\{
\begin{aligned}
&\alpha_{\text{LCB}}(x) &\text{if} \ \ \lambda_{\xi}(x; \theta)|X| < q|\hat{S}| \\
&\epsilon_x, \epsilon_x\sim U[0,1]  &\text{Otherwise}
\end{aligned}
\right.,
\end{equation}
where $\epsilon_x$ is a uniform random variable for reparameterization, and $q$ is a pre-defined quantile of the number of samples. The threshold $q|\hat{S}|$ can be adaptively adjusted during the BO process. To minimize R-LCB, we randomly sample a set of start points and adopt a LBFGS~\citep{lbfgs} optimizer. In particular, LBFGS will not update the samples whose predicted ranking is larger than $q|\hat{S}|$, and they have a probability of being selected as the next query if the sampled $\epsilon_x$ is very small. 
Results in Fig.~\ref{fig:ablation_q} show the advantage of our R-LCB against LCB.

\subsection{Expected Ranking Improvement (ERI)}
Inspired by EI, we introduce ERI to maximize the expected improvement on ranking over the worst tolerable ranking $K_{m}$. We set $K_{m}=5$ by default. 
\begin{align}
\label{eq:eri}
    \alpha_{\text{ERI}}(x) = \sum_{k=0}^{K_{m}} (K_{m} - k) \cdot \text{P}\left( \hat{R}_x=k |\theta, x \right),
\end{align}
where $\text{P}\left( \hat{R}_x=k | \theta, x \right)$ is defined in Eq.~\ref{eq:posterior} representing the prediction of ranking of $x$. The gradient of ERI w.r.t. $x$ is defined as follows, where $\lambda_\xi(x) = \lambda_\xi(x; \theta)$.
\begin{align}
    \frac{\partial\alpha_{\text{ERI}}(x)}{\partial x}
    &=\sum_{k=0}^{K_{m}}\left[
    \frac{K_{m} - k}{k!}\frac{\partial}{\partial x} \left(
        \frac{\left(\lambda_\xi(x)|X|\right)^k}{k!\cdot Z(x)} \cdot
        \exp{\big(-\lambda_\xi(x)|X|\big)}
    \right)
    \right] \\
    &= \sum_{k=0}^{K_{m}} \Bigg\{
    \frac{(K_{m} - k)\left(\lambda_{\xi}(x)|X|\right)^{k-1}|X|}{k!\big( \sum_{i=0}^N\frac{\left(\lambda_{\xi}(x)|X|\right)^2}{i!} \big)^2} \bigg[k \sum_{i=0}^N\frac{\left(\lambda_{\xi}(x)|X|\right)^2}{i!} 
    - \lambda_{\xi}(x)|X| \sum_{i=0}^{N-1}\frac{\left(\lambda_{\xi}(x)|X|\right)^2}{i!}\bigg] \Bigg\}. \notag
\end{align}
Hence, we can get the next query $x^*$ by minimizing $\alpha_{\text{ERI}}(x)$ through a LBFGS optimizer. Similar to R-LCB, we also apply the rectified technique in Eq.~\ref{eq:rLCB} to ERI.

\section{Empirical Analysis} \label{sec:exp}
\begin{wrapfigure}{r}{0.35\linewidth}
\vspace{-15pt}
\centering
\includegraphics[width=0.98\linewidth]{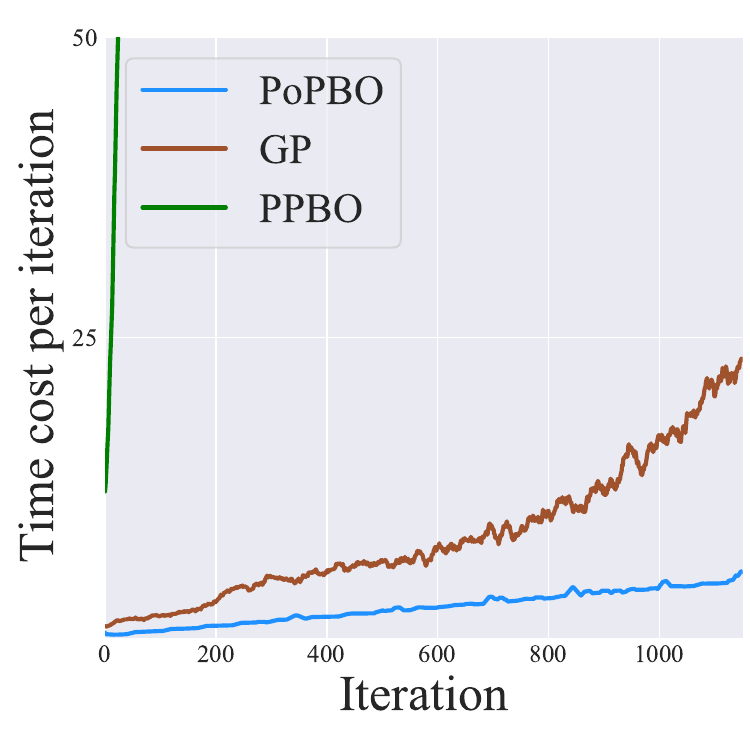} 
\vspace{-18pt}
\caption{Time cost of GP-BO, PPBO~\citep{DBLP:ppbo_conf/icml/MikkolaTJRK20} and PoPBO. All the methods are applied to optimize 6-d Hartmann function. The units are wall-clock times.}
\label{fig:time_cost}
\vspace{-8pt}
\end{wrapfigure}
\textbf{Benchmarks.} We verify the efficacy of PoPBO on both simulated and real-world benchmarks, including HPO and NAS. For the simulated benchmark, we select three simulation functions, including 2-d Branin, 6-d Hartmann, and 6-d Rosenbrock. For the HPO task, we test on the tabular benchmark HPO-Bench~\citep{DBLP:hpobench_conf/nips/EggenspergerMMF21}. For the NAS task, we test on on NAS-Bench-201~\citep{DBLP:nasbench201_conf/iclr/Dong020}. Details of the benchmarks can be found in the Appendix~\ref{supp:sec:benchmark}

\textbf{Baselines.}
We compare against random search (RS)~\citep{DBLP:rs_journals/jmlr/BergstraB12} and various value-based Bayesian optimization methods, including GP~\citep{DBLP:spearmint_conf/nips/SnoekLA12}, TPE~\citep{DBLP:tpe_conf/nips/BergstraBBK11}, SMAC~\citep{DBLP:smac_conf/lion/HutterHL11}, BOHAMIANN~\citep{DBLP:BOHAMIANN_conf/nips/SpringenbergKFH16}, and HEBO~\citep{hebo}. For GP methods, we use EI and LCB as acquisition functions, which are optimized by LBFGS, and adopt the Mat\'ern 5/2 covariance function to be the kernel function. 
We also compare with PPBO~\citep{DBLP:ppbo_conf/icml/MikkolaTJRK20}, one of state of the art preferential BO methods. Detailed settings of the baselines are provided in Appendix~\ref{supp:subsec:baselines}.

\begin{figure*}[tb]
    \centering
    \includegraphics[width=1.0\linewidth]{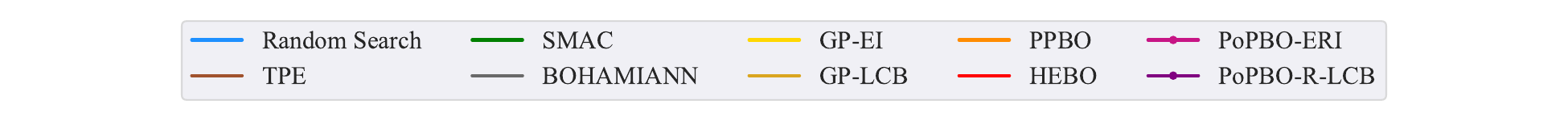}\vspace{-5pt}

  \begin{subfigure}[t]{0.31\linewidth}
    \centering
    \includegraphics[width=0.98\textwidth]{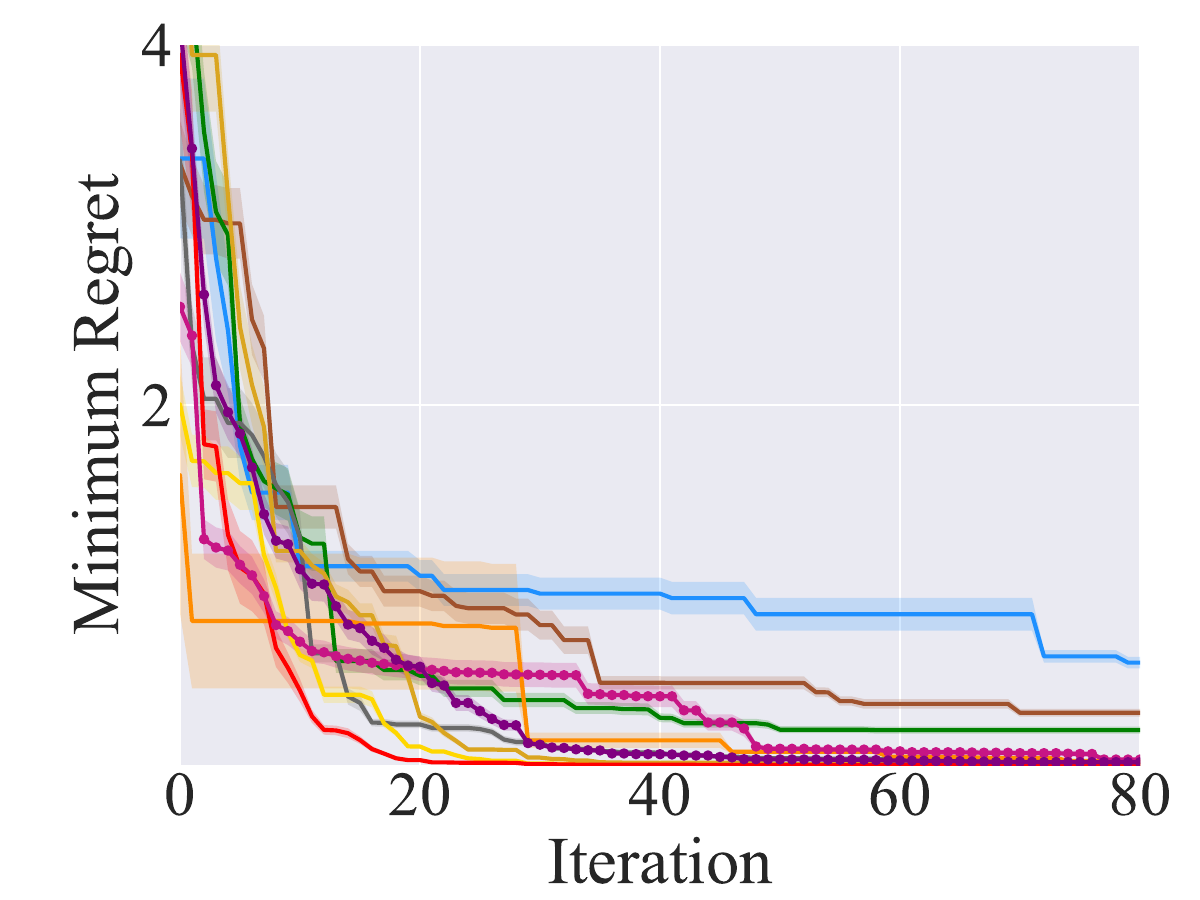}  
    \vspace{-6pt}
    \caption{2-d Branin}
    \label{fig:simulation_branin}
  \end{subfigure}
  \begin{subfigure}[t]{0.31\linewidth}
    \centering
    \includegraphics[width=0.98\textwidth]{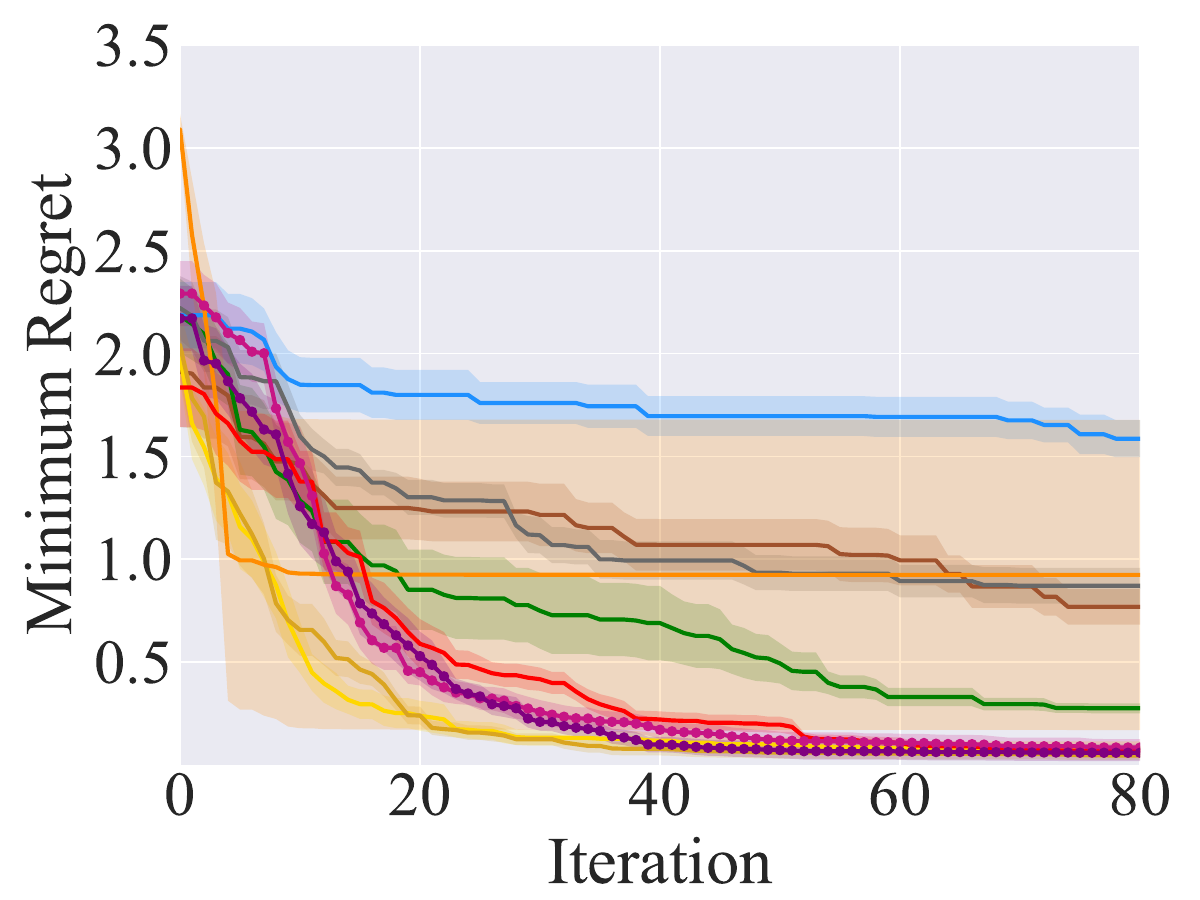}  
    \vspace{-6pt}
    \caption{6-d Hartmann}
    \label{fig:simulation_hartmann}
  \end{subfigure}
  \begin{subfigure}[t]{0.31\linewidth}
    \centering
    \includegraphics[width=0.98\textwidth]{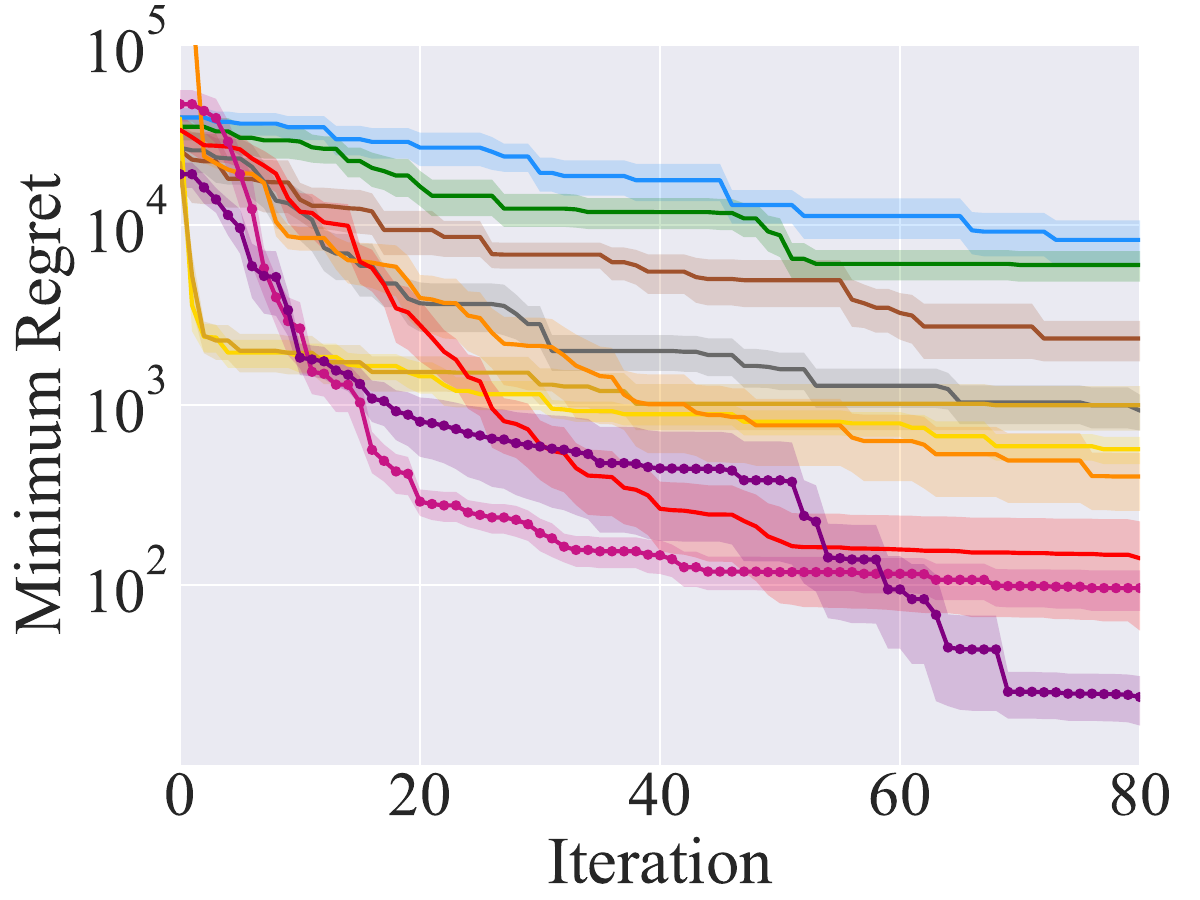}  
    \vspace{-6pt}
    \caption{6-d Rosenbrock}
    \label{fig:simulation_rosenbrock}
  \end{subfigure}
\vspace{-8pt}
\caption{Performance of black-box optimization methods on three simulation functions. Y-axis is the residual of the optimum function value and the incumbent. We run each method ten times and plot the average performance and standard error as the line and shadow.}
\vspace{-10pt}
\label{fig:simulation}
\end{figure*}

\textbf{Settings.}
We run all the methods for 80 iterations with $12$ initial points by default. For the Rosenbrock-6d simulation function, we run all methods for 80 iterations with $30$ initial points due to its complex search space. 
The MLP $\lambda_\xi(x; \theta)$ used to approximate the parameter of the Poisson process has three hidden layers with 128 nodes and a ReLU activation function. The MLP is trained for 100 steps by ADAM~\cite{adam} with 64 batch sizes and a $0.01$ initial learning rate multiplied by 0.2 every 30 steps. All methods are evaluated ten times independently on an Intel(R) Xeon(R) Silver 4210R CPU.

\subsection{Performance on the Simulated Benchmarks}

Fig.~\ref{fig:simulation} compares PoPBO and baselines on 2-d Branin, 6-d Hartmann, and 6-d Rosenbrock simulation benchmarks.
Hartmann has higher dimensions than Branin and thus is more difficult to optimize. As shown in Fig.~\ref{fig:simulation_hartmann}, although the standard GP temporarily outperforms others in the early stage,
our PoPBO achieves the best at around 40-th iterations and takes the lead till the end. In particular, PPBO performs great in the early stage but falls into a local optimum after ten iterations.
Optimizing the Rosebrock function is much more complicated than Branin and Hartmann since its global optimum lies in a narrow valley~\citep{picheny2013benchmark} as well as a more extensive search space. Hence, we increase the initial points to 30 for a better preview of the Rosenbrock landscape for all methods.
Fig.~\ref{fig:simulation_rosenbrock} shows that our PoPBO can quickly find the valley and significantly outperforms other BO methods.

\begin{figure*}[tb]
    \centering
    \includegraphics[width=0.99\linewidth]{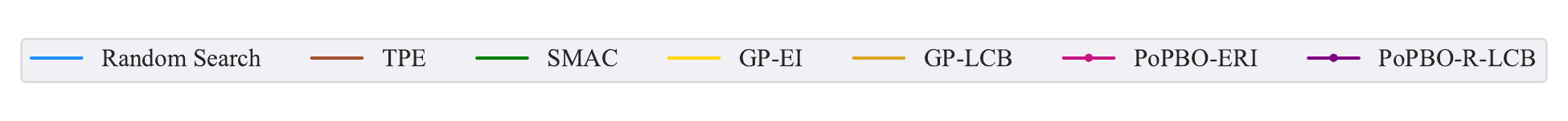}\vspace{-5pt}

  \begin{subfigure}[t]{0.48\linewidth}
    \centering
    \includegraphics[width=0.98\textwidth]{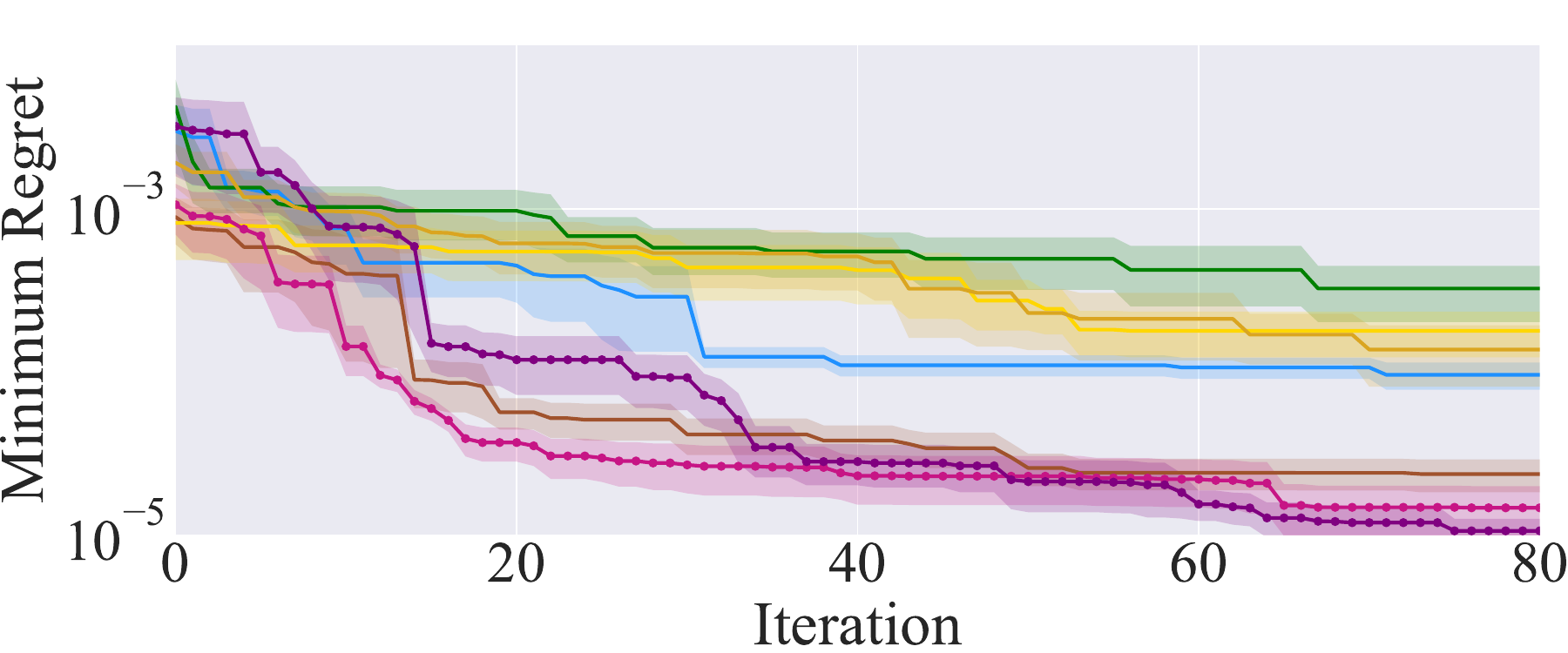}  
    \vspace{-6pt}
    \caption{HPO-Bench-Naval}
    \label{fig:hpobench_naval}
  \end{subfigure}
  \begin{subfigure}[t]{0.48\linewidth}
    \centering
    \includegraphics[width=0.98\textwidth]{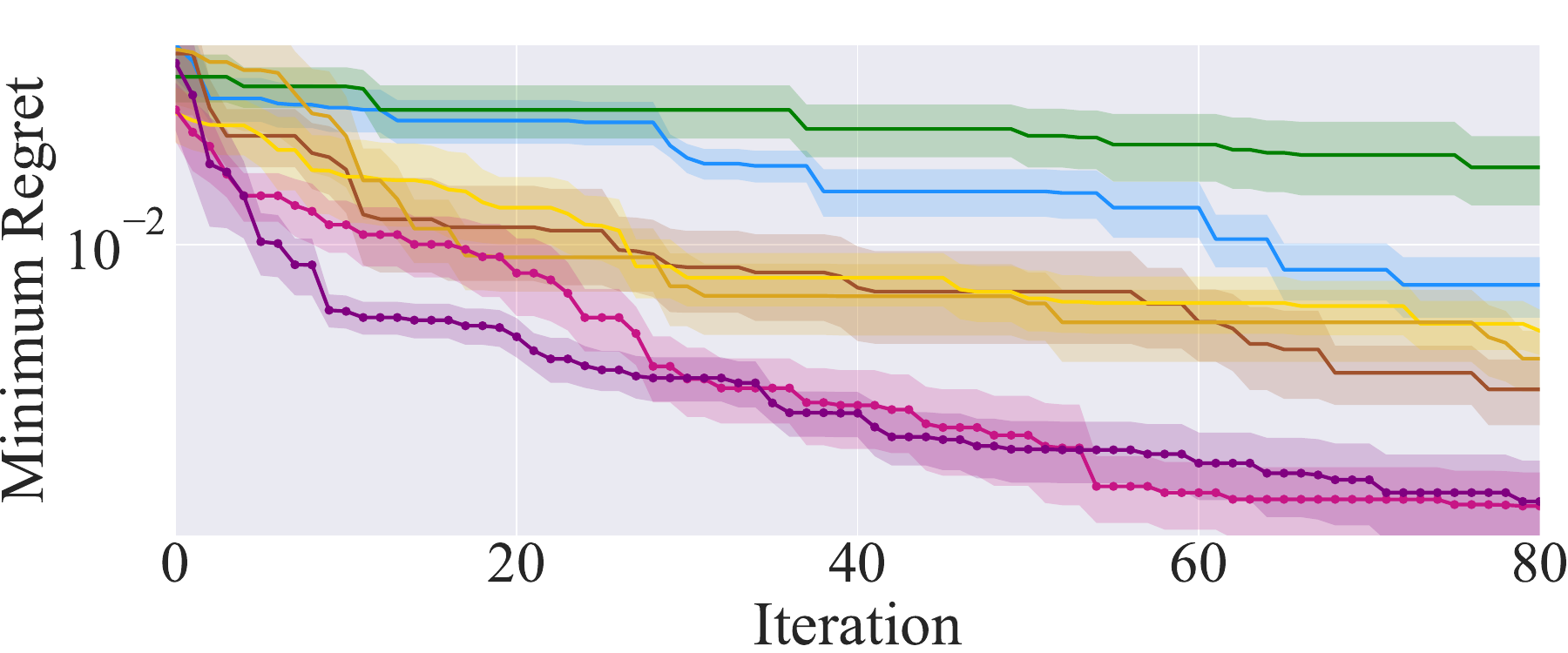}  
    \vspace{-6pt}

    \caption{HPO-Bench-Parkinson}
    \label{fig:hpobench_parkinsons}
  \end{subfigure}
\vspace{-10pt}
  \begin{subfigure}[t]{0.48\linewidth}
    \centering
    \includegraphics[width=0.98\textwidth]{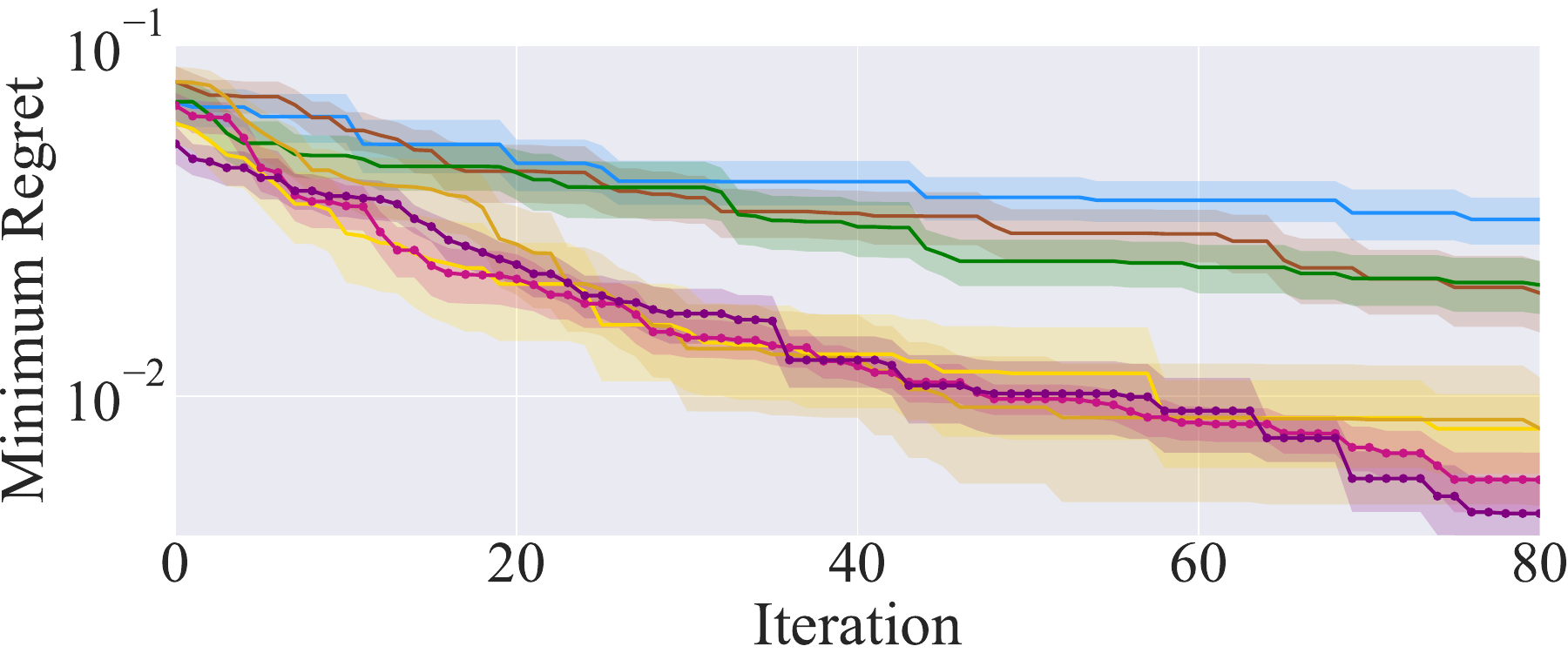}  
    \vspace{-6pt}

    \caption{HPO-Bench-Protein}
    \label{fig:hpobench_protein}
  \end{subfigure}
  \begin{subfigure}[t]{0.48\linewidth}
    \centering
    \includegraphics[width=0.98\textwidth]{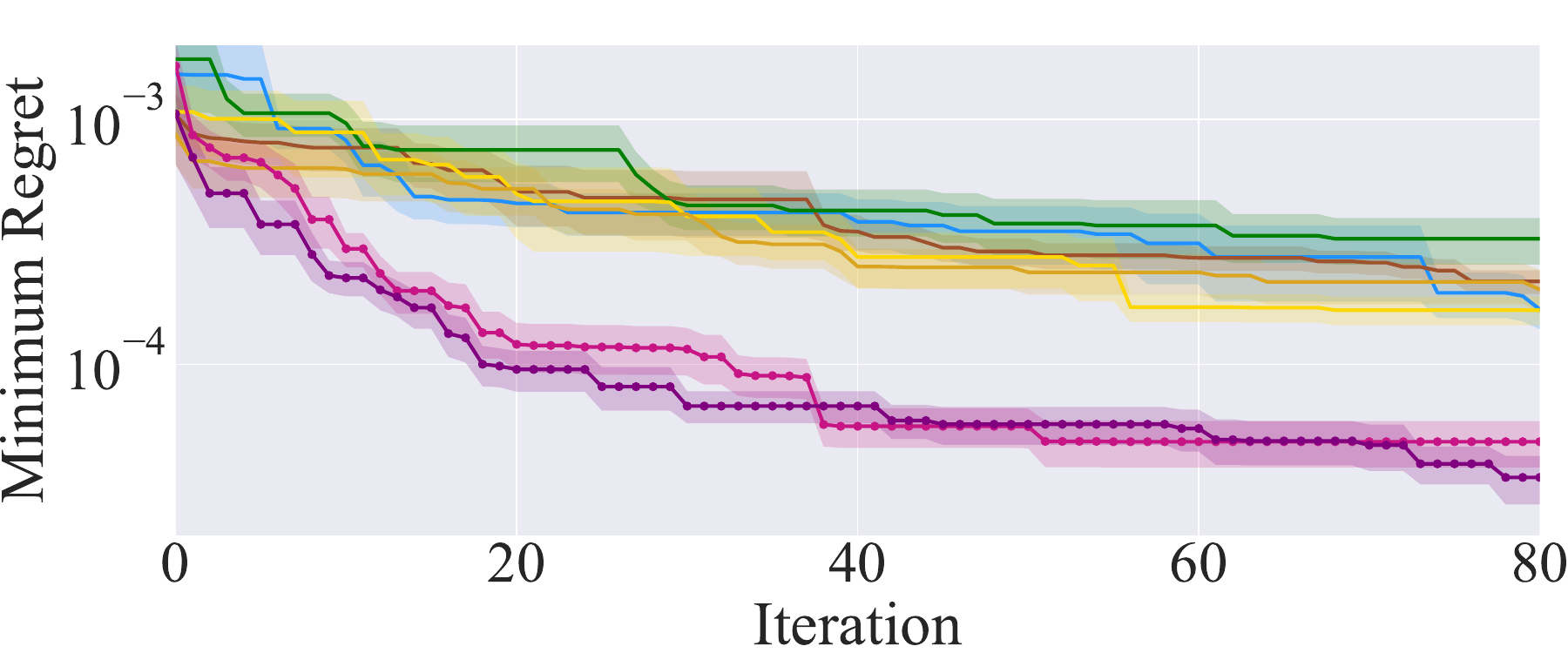}  
    \vspace{-6pt}

    \caption{HPO-Bench-Slice}
    \label{fig:hpobench_slice}
  \end{subfigure}
\caption{Minimum regret comparison with random search and various Bayesian optimization methods on tabular datasets in HPO-Bench. Y-axis indicates the residual between the optimum function value and the incumbent. We run each method ten times and plot the average performance and standard error of the incumbent as the line and shadow. Our PoPBO can quickly discover good samples and achieves the best performance (lowest regret).}
\label{fig:hpobench}
\vspace{-5pt}
\end{figure*}

\textbf{Computational Cost.} Fig.~\ref{fig:time_cost} compares the time cost (wall-clock time) of three peer methods, showing that the cost of GP-BO and PPBO are much higher than PoPBO as the number of observations increases. Specifically, GP has to compute the inverse of a covariance matrix resulting in a $O(N^3)$ computational complexity. PPBO is also based on GP and requires computing another covariance matrix of size $J\times J$, where $J$ is the number of random samples. 
In contrast, the computational bottleneck of PoPBO lies in the training of an MLP, which is $O(N^2)$ as shown in Eq.~\ref{eq:neg_log_likelihood}.
Notice that the units are wall-clock times. Nevertheless, in our experiment, the evaluation of candidates can be negligible compared with the training of the surrogate model and the optimization of acquisition. Therefore, the wall-clock time has little difference from the CPU time.

\subsection{Performance on the Real-world Benchmarks}

\textbf{HPO-Bench.}
we run each method for ten times and plot the trend of minimum regret during the BO procedure. 
Fig.~\ref{fig:hpobench} compares PoPBO with advanced Bayesian optimization methods and random search on HPO-Bench~\citep{DBLP:hpobench_conf/nips/EggenspergerMMF21}, showing that our PoPBO achieves the best on all the four datasets. In contrast, other methods are unable to perform consistently well and even worse than random search. Moreover, the performance of our method has a lower standard error than other methods, indicating its outstanding stability. The numerical performance of all methods on the four datasets are provided in Table~\ref{supp:tab:hpo_bench} in Appendix~\ref{supp:subsec:exp}.

\textbf{NAS-Bench-201.}
Table~\ref{tab:nasbench201} reports the performance on the NAS task. 
The first block shows the performance of prior non-parameter-sharing-based NAS methods, including random search, evolution algorithm, reinforcement learning, and Bayesian optimization. We adopt the same initial observations when testing Random Search, GP-BO, and our PoPBO for a fair comparison. As for SMAC and TPE, we directly run the open-source codes, which have different sampling implementations from ours, making it hard to sample the same initial observations as ours even under the same random seed. 
Our method achieves the best performance on the three datasets and, in particular, outperforms the state-of-the-art Bayesian optimization methods BOHB~\citep{DBLP:bohb_conf/icml/FalknerKH18} and BOHAMIANN~\citep{DBLP:BOHAMIANN_conf/nips/SpringenbergKFH16}. Additionally, we plot the performance trend of various methods on the validation and test set of CIFAR-10, CIFAR-100, and ImageNet16-120 in Fig.~\ref{supp:fig:nas-bench-201} in Appendix~\ref{supp:subsec:exp}. 

\begin{table*}[tb!]
\centering
\caption{Top-1 mean accuracy (\%) for classification on NAS-Bench-201. 
The first block shows the performance of non-parameter sharing algorithms and various Bayesian optimization methods.
The second block shows the performance of PoPBO with ERI and R-LCB acquisition functions. $^\dagger$: Results are obtained from NAS-Bench-201. Otherwise, we independently run the method for ten times. The best mean accuracy in each column is in bold.
}
\vspace{-8pt}
\setlength{\tabcolsep}{2pt}
\resizebox{.85\textwidth}{!}{
\smallskip\begin{tabular}{lcccccc}
\toprule
\multirow{2}{*}{\textbf{Methods}} & \multicolumn{2}{c}{\textbf{CIFAR-10}}&  \multicolumn{2}{c}{\textbf{CIFAR-100}} & \multicolumn{2}{c}{\textbf{ImageNet-16-120}}  \\
 \cmidrule(lr){2-3}
 \cmidrule(lr){4-5}
 \cmidrule(lr){6-7}
 & \textbf{valid} & \textbf{test} &  \textbf{valid} & \textbf{test} & \textbf{valid} & \textbf{test} \\
\midrule
REINFORCE~\citep{DBLP:reinforce_journals/ml/Williams92}$^\dagger$  &
91.09$\pm$0.37 &
93.85$\pm$0.37 &
71.61$\pm$1.12 &
71.71$\pm$1.09 &
45.05$\pm$1.02 &
45.24$\pm$1.18 \\
REA~\citep{DBLP:rea_conf/aaai/RealAHL19}$^\dagger$  &
91.19$\pm$0.31 &
93.92$\pm$0.30 &
71.81$\pm$1.12 &
71.84$\pm$0.99 &
45.15$\pm$0.89 &
45.54$\pm$1.03 \\
Random Search~\citep{DBLP:rs_journals/jmlr/BergstraB12}  &
91.00$\pm$0.38 &
93.83$\pm$0.31 &
71.29$\pm$1.29 &
71.47$\pm$1.16 &
44.83$\pm$1.11 &
45.05$\pm$1.14 \\
BOHB~\citep{DBLP:bohb_conf/icml/FalknerKH18}$^\dagger$  &
90.82$\pm$0.53 &
93.61$\pm$0.52 &
70.74$\pm$1.29 &
70.85$\pm$1.28 &
44.26$\pm$1.36 &
44.42$\pm$1.49 \\
TPE~\citep{DBLP:tpe_conf/nips/BergstraBBK11} & 91.06$\pm$0.38 & 93.90$\pm$0.34 & 71.29$\pm$1.29 & 71.85$\pm$1.13 & 45.04$\pm$1.23 & 45.27$\pm$1.49 \\
SMAC~\citep{DBLP:smac_conf/lion/HutterHL11} & 91.09$\pm$0.38 & 93.95$\pm$0.28 & 71.40$\pm$1.23 & 71.66$\pm$1.11 & 45.11$\pm$0.99 & 45.32$\pm$1.09 \\
GP (EI)~\citep{DBLP:spearmint_conf/nips/SnoekLA12} & 91.40$\pm$0.18 & 94.23$\pm$0.15 & 72.67$\pm$0.83 & 72.75$\pm$0.47 & 45.83$\pm$0.45 & 46.20$\pm$0.63 \\
GP (LCB)~\citep{DBLP:spearmint_conf/nips/SnoekLA12} & 91.30$\pm$0.25 & 93.98$\pm$0.22 & 72.00$\pm$0.80 & 72.05$\pm$0.76 & 45.37$\pm$0.83 & 45.60$\pm$0.94 \\
BOHAMIANN~\citep{DBLP:BOHAMIANN_conf/nips/SpringenbergKFH16} & 91.36$\pm$0.16 & 94.13$\pm$0.23 & 72.36$\pm$0.82 & 72.38$\pm$0.81 & 45.93$\pm$0.66 & 46.18$\pm$0.60 \\
\midrule
PoPBO (ERI) & \textbf{91.52$\pm$0.05} & \textbf{94.35$\pm$0.03} & \textbf{73.21$\pm$0.29} & \textbf{73.25$\pm$0.18} & \textbf{46.27$\pm$0.36} & 46.54$\pm$0.19 \\
PoPBO (R-LCB) & \textbf{91.52$\pm$0.04} & 94.33$\pm$0.08 & \textbf{73.21$\pm$0.36} & 73.19$\pm$0.31 & 46.12$\pm$0.43 & \textbf{46.61$\pm$0.32} \\
\bottomrule
\end{tabular}
}
\label{tab:nasbench201}
\vspace{-3pt}
\end{table*}

\subsection{Effectiveness of the Rectified Technique}

\begin{figure*}[tb]
    \centering
  \begin{subfigure}[t]{0.48\linewidth}
    \centering
    \includegraphics[width=0.98\textwidth]{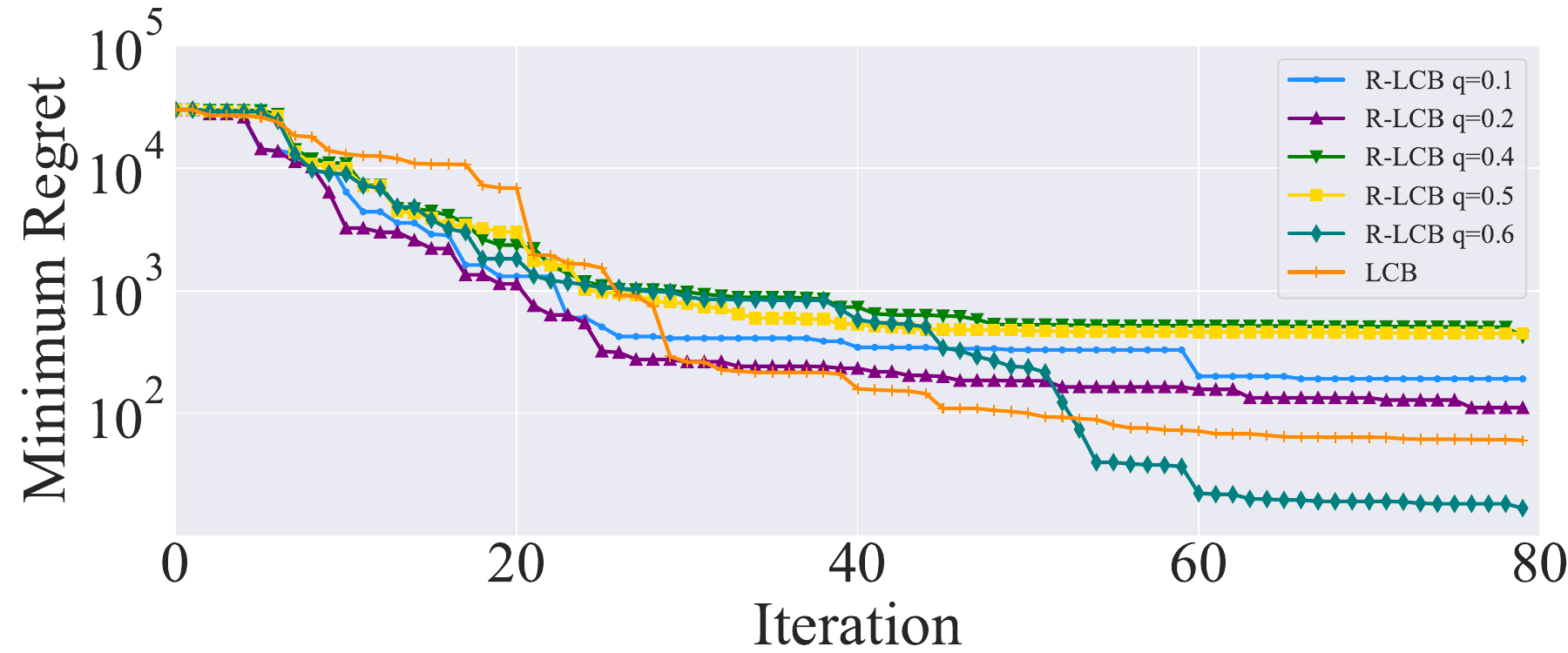}  
    \vspace{-6pt}
    \caption{Results of R-LCB on Rosenbrock}
    \label{fig:ab_rlcb_rosenbrock}
  \end{subfigure}
  \begin{subfigure}[t]{0.48\linewidth}
    \centering
    \includegraphics[width=0.98\textwidth]{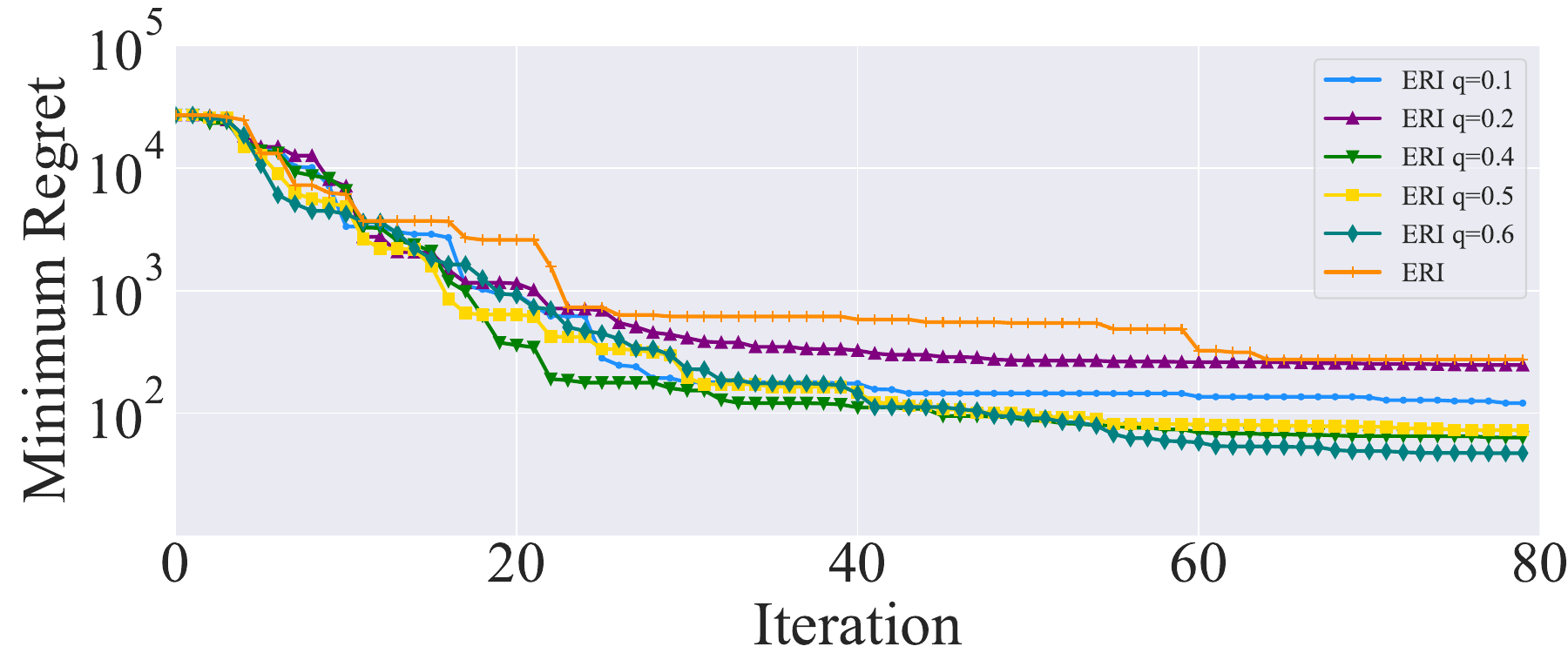}  
    \vspace{-6pt}
    \caption{Results of ERI on Rosenbrock}
    \label{fig:ab_eri_rosenbrock}
  \end{subfigure}
\vspace{-8pt}
\caption{Ablation study on hyperparameter $q$, which controls the exploitation-exploration trade-off. We test on the Rosenbrock simulation function and show the effect of $q$ on (a) R-LCB and (b) ERI. For each setting, we run ten times and plot the average performance of the incumbent.}
\vspace{-5pt}
\label{fig:ablation_q}
\end{figure*}

The quantile parameter $q$ in Eq.~\ref{eq:rLCB} trades off the exploration and exploitation, i.e., a smaller $q$ has better exploration ability. 
On the one hand, the method degrades to random search when $q\rightarrow0$, with weak exploitation ability, so it is undesirable to set $q$ as a rather small value. 
On the other hand, it degrades to the vanilla acquisition functions when $q\rightarrow1$, with weak exploration ability as our analysis in Sec.~\ref{sec:acq_func}. 
Fig.~\ref{fig:ablation_q} evaluates the effect of quantile parameters $q$ on both R-LCB and ERI on 6-d Rosenbrock benchmark.
The best setting of the quantile parameter $q$ for R-LCB is $0.6$, and the best for ERI is $0.4$. We also conduct the ablation study on NAS-Bench-201, which verify that this setting could provide a good balance between exploration and exploitation. The results on NAS-Bench-201 can be found in Appendix~\ref{supp:sub:nasbench_ab}.

\section{Conclusion} \label{sec:conclusion}
We have proposed a novel Bayesian Optimization framework, named PoPBO, for optimizing black-box functions with relative responses, being more robust to noise than absolute responses. Specifically, we introduce a relative response surface to capture the global ranking of candidates based on the Poisson process that is suitable for modeling discrete count events. We give the likelihood and posterior forms of ranking under the general assumption of a non-homogeneous Poisson process. To balance the trade-off between exploration and exploitation, we design two acquisition functions, namely Rectified Lower Confidence Bound (R-LCB) and Expected Ranking Improvement (ERI), for our ranking-based response surface. 
Our method enjoys a lower computational complexity of $O(N^2)$ compared to GP's $O(N^3)$ and performs competitively on both simulated and real-world benchmarks.

\textbf{Limitations and Future Work.} 
This work analyzes the robustness of relative response against noise and thus does not involve prior knowledge of noise. However, there exist real scenarios where the noise is too large to disrupt the ranking of observations, and we would like to leave it as our future work.
Additionally, the mean of Poisson distribution is the same as variance, which has a potential over-exploitation issue as mentioned in Sec.~\ref{sec:acq_func}. This work introduces a rectified technique to alleviate it, and we would like to explore other elegant acquisition functions in future work.
\section*{Acknowledgements}
This work was supported by NSFC (U19B2035, 61972250, 61671480), the Qingdao Natural Science Foundation (23-2-1-161-zyyd-jch), and the Shandong Natural Science Foundation (ZR2023MF008).


\bibliography{PoPBO}

\begin{thebibliography}{}

\bibitem[Agrawal and Goyal, 2013]{DBLP:thompson_conf/icml/AgrawalG13}
Agrawal, S. and Goyal, N. (2013).
\newblock Thompson sampling for contextual bandits with linear payoffs.
\newblock In {\em ICML}.

\bibitem[Bergstra et~al., 2011]{DBLP:tpe_conf/nips/BergstraBBK11}
Bergstra, J., Bardenet, R., Bengio, Y., and K{\'{e}}gl, B. (2011).
\newblock Algorithms for hyper-parameter optimization.
\newblock In {\em NeurIPS}.

\bibitem[Bergstra and Bengio, 2012]{DBLP:rs_journals/jmlr/BergstraB12}
Bergstra, J. and Bengio, Y. (2012).
\newblock Random search for hyper-parameter optimization.
\newblock {\em Journal Of Machine Learning Research}.

\bibitem[Brochu et~al., 2010]{DBLP:no_closed_func}
Brochu, E., Cora, V.~M., and de~Freitas, N. (2010).
\newblock A tutorial on bayesian optimization of expensive cost functions, with
  application to active user modeling and hierarchical reinforcement learning.
\newblock {\em Computer Research Repository}.

\bibitem[Brusilovsky et~al., 2007]{DBLP:noabsolute_conf/adaptive/2007}
Brusilovsky, P., Kobsa, A., and Nejdl, W., editors (2007).
\newblock {\em The Adaptive Web, Methods and Strategies of Web
  Personalization}, volume 4321 of {\em Lecture Notes in Computer Science}.
  Springer.

\bibitem[Calandra et~al., 2016]{DBLP:robotic_journals/amai/CalandraSPD16}
Calandra, R., Seyfarth, A., Peters, J., and Deisenroth, M.~P. (2016).
\newblock Bayesian optimization for learning gaits under uncertainty - an
  experimental comparison on a dynamic bipedal walker.
\newblock {\em Annals of Mathematics and Artificial Intelligence}.

\bibitem[Chapelle and Li, 2011]{DBLP:thompson_conf/nips/ChapelleL11}
Chapelle, O. and Li, L. (2011).
\newblock An empirical evaluation of thompson sampling.
\newblock In {\em NeurIPS}.

\bibitem[Chen et~al., 2014]{DBLP:sghmc_conf/icml/ChenFG14}
Chen, T., Fox, E.~B., and Guestrin, C. (2014).
\newblock Stochastic gradient hamiltonian monte carlo.
\newblock In {\em ICML}.

\bibitem[Coraddu et~al., 2016]{naval_coraddu2016machine}
Coraddu, A., Oneto, L., Ghio, A., Savio, S., Anguita, D., and Figari, M.
  (2016).
\newblock Machine learning approaches for improving condition-based maintenance
  of naval propulsion plants.
\newblock {\em Proceedings of the Institution of Mechanical Engineers, Part M:
  Journal of Engineering for the Maritime Environment}.

\bibitem[Cowen{-}Rivers et~al., 2022]{hebo}
Cowen{-}Rivers, A.~I., Lyu, W., Wang, Z., Tutunov, R., Hao, J., Wang, J., and
  Bou{-}Ammar, H. (2022).
\newblock Hebo: Pushing the limits of sample-efficient hyper-parameter
  optimisation.
\newblock {\em Journal of Artificial Intelligence Research}.

\bibitem[Dong and Yang, 2020]{DBLP:nasbench201_conf/iclr/Dong020}
Dong, X. and Yang, Y. (2020).
\newblock Nas-bench-201: Extending the scope of reproducible neural
  architecture search.
\newblock In {\em ICLR}.

\bibitem[Eggensperger et~al., 2021]{DBLP:hpobench_conf/nips/EggenspergerMMF21}
Eggensperger, K., M{\"{u}}ller, P., Mallik, N., Feurer, M., Sass, R., Klein,
  A., Awad, N.~H., Lindauer, M., and Hutter, F. (2021).
\newblock Hpobench: {A} collection of reproducible multi-fidelity benchmark
  problems for {HPO}.
\newblock In {\em NeurIPS}.

\bibitem[Falkner et~al., 2018]{DBLP:bohb_conf/icml/FalknerKH18}
Falkner, S., Klein, A., and Hutter, F. (2018).
\newblock {BOHB:} robust and efficient hyperparameter optimization at scale.
\newblock In {\em ICML}.

\bibitem[Feurer et~al., 2018]{transfer_feurer2018scalable}
Feurer, M., Letham, B., and Bakshy, E. (2018).
\newblock Scalable meta-learning for bayesian optimization using
  ranking-weighted gaussian process ensembles.
\newblock In {\em AutoML Workshop at ICML}.

\bibitem[Gonz{\'{a}}lez et~al., 2017]{DBLP:PBO_conf/icml/GonzalezDDL17}
Gonz{\'{a}}lez, J., Dai, Z., Damianou, A.~C., and Lawrence, N.~D. (2017).
\newblock Preferential bayesian optimization.
\newblock In {\em ICML}.

\bibitem[González et~al., 2015]{bo_bio_genedesign}
González, J., Longworth, J., James, D.~C., and Lawrence, N.~D. (2015).
\newblock Bayesian optimization for synthetic gene design.
\newblock {\em arXiv preprint arXiv:1505.01627}.

\bibitem[Graf et~al., 2011]{DBLP:slice_conf/miccai/GrafKSPC11}
Graf, F., Kriegel, H., Schubert, M., P{\"{o}}lsterl, S., and Cavallaro, A.
  (2011).
\newblock 2d image registration in {CT} images using radial image descriptors.
\newblock In {\em MICCAI}.

\bibitem[He et~al., 2022]{gnnrank}
He, Y., Gan, Q., Wipf, D., Reinert, G., Yan, J., and Cucuringu, M. (2022).
\newblock Gnnrank: Learning global rankings from pairwise comparisons via
  directed graph neural networks.
\newblock In {\em ICML}.

\bibitem[Hutter et~al., 2011]{DBLP:smac_conf/lion/HutterHL11}
Hutter, F., Hoos, H.~H., and Leyton{-}Brown, K. (2011).
\newblock Sequential model-based optimization for general algorithm
  configuration.
\newblock In {\em LION}.

\bibitem[Kahneman and Tversky, 2013]{noabsolute_kahneman2013prospect}
Kahneman, D. and Tversky, A. (2013).
\newblock Prospect theory: An analysis of decision under risk.
\newblock In {\em Handbook of the fundamentals of financial decision making:
  Part I}, pages 99--127. World Scientific.

\bibitem[Klein and Hutter, 2019]{DBLP:fcnet_journals/corr/abs-1905-04970}
Klein, A. and Hutter, F. (2019).
\newblock Tabular benchmarks for joint architecture and hyperparameter
  optimization.
\newblock {\em Computer Research Repository}.

\bibitem[Lin et~al., 2022]{bope}
Lin, Z.~J., Astudillo, R., Frazier, P.~I., and Bakshy, E. (2022).
\newblock Preference exploration for efficient bayesian optimization with
  multiple outcomes.
\newblock In {\em AISTATS}, volume 151, pages 4235--4258.

\bibitem[Liu and Nocedal, 1989]{lbfgs}
Liu, D.~C. and Nocedal, J. (1989).
\newblock On the limited memory {BFGS} method for large scale optimization.
\newblock {\em Math. Program.}, 45(1-3):503--528.

\bibitem[Mikkola et~al., 2020]{DBLP:ppbo_conf/icml/MikkolaTJRK20}
Mikkola, P., Todorovic, M., J{\"{a}}rvi, J., Rinke, P., and Kaski, S. (2020).
\newblock Projective preferential bayesian optimization.
\newblock In {\em ICML}.

\bibitem[Mockus, 1994]{DBLP:ei_journals/jgo/Mockus94}
Mockus, J. (1994).
\newblock Application of bayesian approach to numerical methods of global and
  stochastic optimization.
\newblock {\em Journal of Global Optimization}.

\bibitem[Mockus et~al., 1978]{mockus1978application}
Mockus, J., Tiesis, V., and Zilinskas, A. (1978).
\newblock The application of bayesian methods for seeking the extremum.
\newblock {\em Towards global optimization}.

\bibitem[Nguyen et~al., 2021]{DBLP:topk_conf/aaai/NguyenTLJ21}
Nguyen, Q.~P., Tay, S., Low, B. K.~H., and Jaillet, P. (2021).
\newblock Top-k ranking bayesian optimization.
\newblock In {\em AAAI}.

\bibitem[Perrone et~al., 2018]{DBLP:ablr}
Perrone, V., Jenatton, R., Seeger, M.~W., and Archambeau, C. (2018).
\newblock Scalable hyperparameter transfer learning.
\newblock In {\em NeurIPS}.

\bibitem[Picheny et~al., 2013]{picheny2013benchmark}
Picheny, V., Wagner, T., and Ginsbourger, D. (2013).
\newblock A benchmark of kriging-based infill criteria for noisy optimization.
\newblock {\em Structural and multidisciplinary optimization}.

\bibitem[Rana, 2013]{protein_rana2013physicochemical}
Rana, P. (2013).
\newblock Physicochemical properties of protein tertiary structure data set.
\newblock {\em UCI Machine Learning Repository}.

\bibitem[Real et~al., 2019]{DBLP:rea_conf/aaai/RealAHL19}
Real, E., Aggarwal, A., Huang, Y., and Le, Q.~V. (2019).
\newblock Regularized evolution for image classifier architecture search.
\newblock In {\em AAAI}.

\bibitem[Rosset et~al., 2005]{RankingICDM05}
Rosset, S., Perlich, C., and Zadrozny, B. (2005).
\newblock Ranking-based evaluation of regression models.
\newblock In {\em ICDM}.

\bibitem[Salinas et~al., 2020]{DBLP:shen_conf/icml/SalinasSP20}
Salinas, D., Shen, H., and Perrone, V. (2020).
\newblock A quantile-based approach for hyperparameter transfer learning.
\newblock In {\em ICML}.

\bibitem[Siivola et~al., 2021]{DBLP:PBBO_conf/mlsp/SiivolaDAGMV21}
Siivola, E., Dhaka, A.~K., Andersen, M.~R., Gonz{\'{a}}lez, J., Moreno, P.~G.,
  and Vehtari, A. (2021).
\newblock Preferential batch bayesian optimization.
\newblock In {\em MLSP}.

\bibitem[Snoek et~al., 2012]{DBLP:spearmint_conf/nips/SnoekLA12}
Snoek, J., Larochelle, H., and Adams, R.~P. (2012).
\newblock Practical bayesian optimization of machine learning algorithms.
\newblock In {\em NeurIPS}.

\bibitem[Snoek et~al., 2015]{DBLP:dngo_conf/icml/SnoekRSKSSPPA15}
Snoek, J., Rippel, O., Swersky, K., Kiros, R., Satish, N., Sundaram, N.,
  Patwary, M. M.~A., Prabhat, and Adams, R.~P. (2015).
\newblock Scalable bayesian optimization using deep neural networks.
\newblock In {\em ICML}.

\bibitem[Song et~al., 2022]{likelihoodfree}
Song, J., Yu, L., Neiswanger, W., and Ermon, S. (2022).
\newblock A general recipe for likelihood-free bayesian optimization.
\newblock In {\em ICML}, volume 162, pages 20384--20404. {PMLR}.

\bibitem[Springenberg et~al., 2016]{DBLP:BOHAMIANN_conf/nips/SpringenbergKFH16}
Springenberg, J.~T., Klein, A., Falkner, S., and Hutter, F. (2016).
\newblock Bayesian optimization with robust bayesian neural networks.
\newblock In {\em NeurIPS}.

\bibitem[Srinivas et~al., 2012]{DBLP:ucb_journals/tit/SrinivasKKS12}
Srinivas, N., Krause, A., Kakade, S.~M., and Seeger, M.~W. (2012).
\newblock Information-theoretic regret bounds for gaussian process optimization
  in the bandit setting.
\newblock {\em IEEE Transactions on Information Theory}.

\bibitem[Tian and Parikh, 2022]{adam}
Tian, R. and Parikh, A.~P. (2022).
\newblock Amos: An adam-style optimizer with adaptive weight decay towards
  model-oriented scale.
\newblock {\em CoRR}, abs/2210.11693.

\bibitem[Tiao et~al., 2021]{bore}
Tiao, L.~C., Klein, A., Seeger, M.~W., Bonilla, E.~V., Archambeau, C., and
  Ramos, F. (2021).
\newblock {BORE:} bayesian optimization by density-ratio estimation.
\newblock In {\em ICML}, volume 139, pages 10289--10300. {PMLR}.

\bibitem[Tsanas et~al., 2010]{DBLP:parkinson_journals/tbe/TsanasLMR10}
Tsanas, A., Little, M.~A., McSharry, P.~E., and Ramig, L.~O. (2010).
\newblock Accurate telemonitoring of parkinson's disease progression by
  noninvasive speech tests.
\newblock {\em IEEE Transactions on Biomedical Engineering}.

\bibitem[Williams, 1992]{DBLP:reinforce_journals/ml/Williams92}
Williams, R.~J. (1992).
\newblock Simple statistical gradient-following algorithms for connectionist
  reinforcement learning.
\newblock {\em Machine Learning}.

\bibitem[Yigiter and Inal, 2006]{yigiter2006right}
Yigiter, A. and Inal, C. (2006).
\newblock Right truncated homogeneous poisson process.
\newblock {\em PAKISTAN JOURNAL OF STATISTICS-ALL SERIES-}, 22(1):69.

\end{thebibliography}





\appendix

\section{Discussion on the Assumptions for $\hat{R}_x(S)$}\label{supp:sec:assumption_discussion}
Consider to minimize a black-box function $f(x)$. In this paper, $\hat{R}_x(S) = |S_x\cap \hat{S}|$, where $\hat{S}$ is a set of samples and $S_x=\{y| y\in S, f(y)<f(x)\}$ is the set of better points than $x$. We assume $\hat{R}_x(S)$ has the following properties: 

1) $\hat{R}_x(S_1) \perp \!\!\! \perp
\hat{R}_x(S_2), \forall S_1,S_2\subset X, S_1\cap S_2 = \emptyset$; (independent increment)

2) $\hat{R}_x(\emptyset) = 0$; 

3) $\lim_{\Delta s\rightarrow0}\text{P}(\hat{R}_x(S')+\Delta s) - \hat{R}_x(S')\geq2) = 0, \forall S'\subset S$. 

We adopt the first assumption since $f(x)$ is a black-box function. Unlike GP that assume $f(x)$ is a linear function $f(x) = w^\top x$, we make no prior assumption on the form of black-box function and only assume the rankings over disjoint areas are independent.
The second assumption is naturally satisfied since there are no points $x'\in \emptyset$ satisfying $y' > y$, where $y$ indicates the observation of point $x$. 
The third assumption is also satisfied no matter if the black-box function is a discrete or continuous. Specifically, we can find a small enough $\Delta s$ making $(S\cap \hat{S})\cap \Delta s$ only contains one point, since $\hat{S}$ is discrete.

\section{Algorithm Details}~\label{supp:sec:alg}

As the number of observations $N$ increases, the right truncated Poisson distribution gradually approaches to the normal one\citep{yigiter2006right}, which has a smaller computational cost. Hence, we use normal Poisson process to model the response surface when $N\geq12$.

\subsection{Workflow of PoPBO}

Suppose at $t$-th iteration, we have $N^{(t)}$ history queries, denoted as $\hat{S}^{(t)}$ and their observations. The detailed workflow of our PoPBO at $t$ iteration is as follows (In the following, we omit $t$ without loss of generality.):

\textbf{1) Get the rankings.} $\forall x\in \hat{S}$, we compute the ranking of $x$ over $\hat{S}$ by comparing its observation with others and obtain a ranking set $\hat{K} = \{k_x | \forall x\in \hat{S}\}$. Similar to [1], we assume the rankings between queries are independent. (Though it is not rigorous, we find it performs better than a truly independent ranking strategy, which will be analyzed later.)

\textbf{2) Compute log-likelihood on observations $\hat{K}$.} Since the rankings in $\hat{K}$ are independent, we can get the log-likelihood on the observations of ranking $\hat{K}$ as:
\begin{align}
    &\log{L(\hat{K}|\hat{S}; \theta)} = \log\Bigg\{\prod_{x\in \hat{S}} p(\hat{R}_{x}=k_x|x,\hat{S})\Bigg\} \quad \text{(Property of independent increment)} \\
    & = \sum_{x\in \hat{S}} \log\Bigg\{ \frac{\left(\lambda(\xi, x)|X|\right)^k}{k!\cdot Z(x)} \exp{\left(-\lambda(\xi, x)|X|\right)}\Bigg\} \quad \text{(According to the definition of $p(\hat{R}_x=k_x|x, \hat{S})$ in Eq.~\ref{eq:norm_prob})} \\
    &=\sum_{x\in \hat{S}}\Bigg\{ k_{x} \log{\left(\lambda_\xi(x; \theta)|X| \right)} - \log{(k_x!)} - \log{\bigg[ \sum_{i=0}^{N-1}\frac{\left(\lambda_\xi(x; \theta)|X| \right)^i}{i!} \bigg]} \Bigg\} \quad \text{(Derive Eq.~\ref{eq:neg_log_likelihood})}. 
\end{align}

\textbf{3) Train the surrogate model.} We then train MLP by minimizing the log-likelihood through ADAM, whose gradient can be computed as Eq. 8.

\textbf{4) Get the next query.} We utilize R-LCB or ERI acquisition function to determine the next query. After obtaining the reward, we put the new query and its reward to the observation set $\hat{S}$.

\subsection{Independent ranking strategy VS. Our implementation.}
Given two points $x_1, x_2$ and a set $S$, their rankings over $S\backslash\{x_1, x_2\}$ are independent. Hence, we implement a truly independent ranking strategy as follows: 
We sample $N/2$ samples from all queries $\hat{S}$ to build a base set $B$. For the left $N/2$ samples, we obtain their rankings over $B$ and build the ranking set $\hat{K} = \{k_x, \forall x \in \hat{S} \backslash B\}$, where $k_x$ denotes the ranking of $x$ over $B$. Such a ranking strategy can guarantee that the rankings in $\hat{K}$ are independent of each other. Then the log-likelihood of the observations of ranking $\hat{K}$ can be computed as:
\begin{align}
    &\log{L(\hat{K}|\hat{S}\backslash B; \theta)} = \log\Bigg\{\prod_{x\in \hat{S}\backslash B} p(\hat{R}_{x}=k_x)\Bigg\} \\
    & = \sum_{x\in \hat{S}\backslash B} \log\Bigg\{ \frac{\left(\lambda(\xi, x)|X|\right)^k}{k!\cdot Z(x)} \exp{\left(-\lambda(\xi, x)|X|\right)}\Bigg\} \notag\\
    &=\sum_{x\in \hat{S}\backslash B}\Bigg\{ k_{x} \log{\left(\lambda_\xi(x; \theta)|X| \right)} - \log{(k_x!)} - \log{\bigg[ \sum_{i=0}^{|B|-1}\frac{\left(\lambda_\xi(x; \theta)|X| \right)^i}{i!} \bigg]} \Bigg\} \notag
\end{align}

However, the above independent ranking strategy results in under-utilization of observations $\hat{S}$ since it has to be divided into two sets. Moreover, candidates with different performances may get the same ranking over $B$, which will mislead the training of the surrogate. This work adopts a simple relaxation -- We compute $\hat{R}$ as the ranking over the whole set $\hat{S}$ and assume independence property, which is also utilized in [1]. Our experimental results show that such a relaxation results in better performance since it can fully utilize all observations.

\begin{figure*}[tb!]
    \centering
    \includegraphics[width=1.0\textwidth]{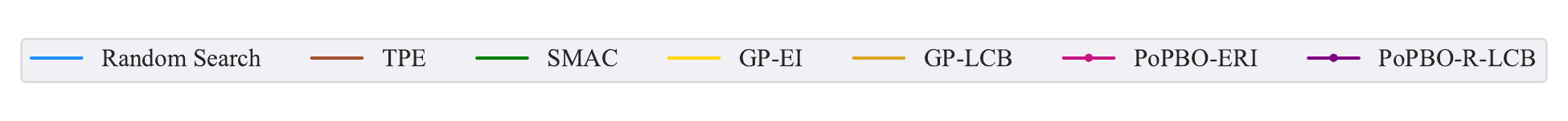}\vskip -5pt
  \begin{subfigure}[t]{0.31\linewidth}
    \centering
    \includegraphics[width=0.98\textwidth]{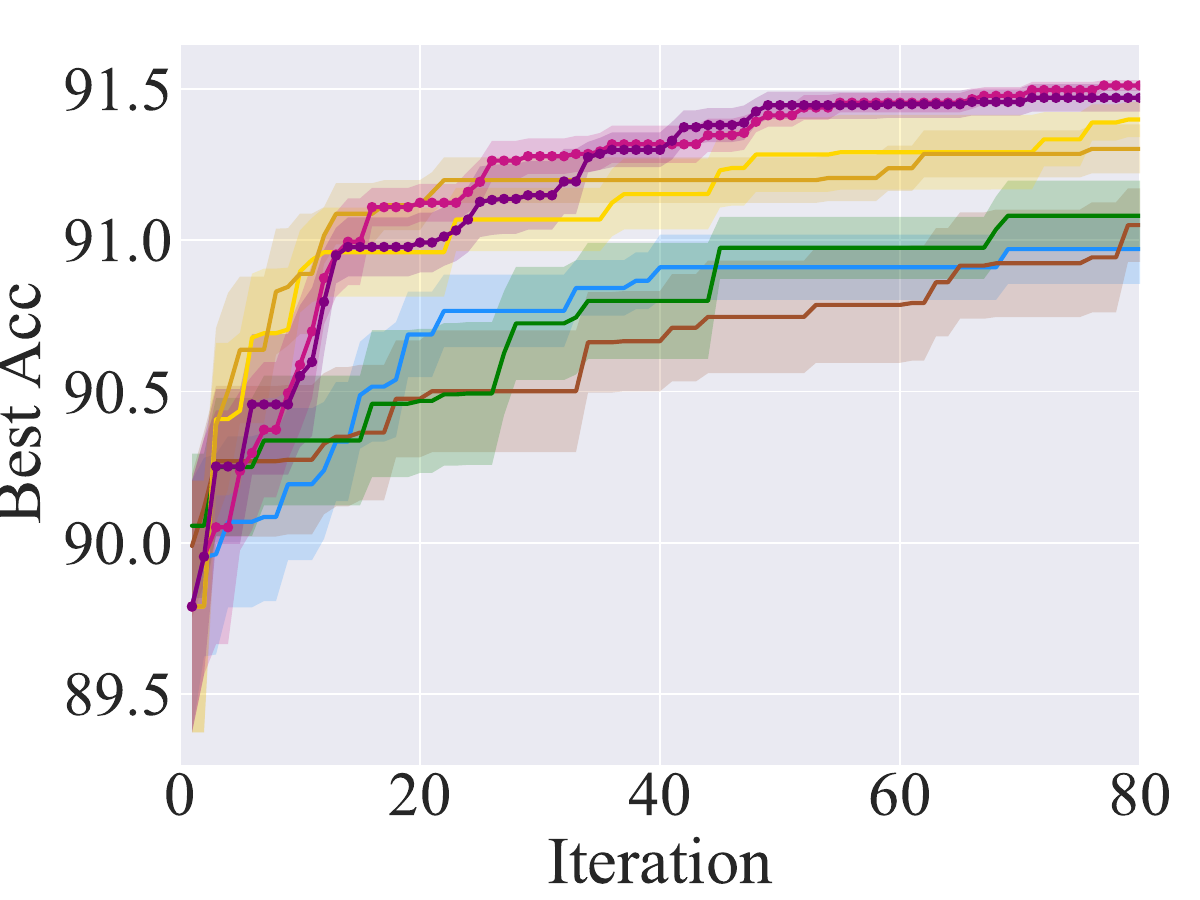}  
    \caption{CIFAR-10 valid-set}
    \label{supp:fig:cifar10_valid}
  \end{subfigure}
  \begin{subfigure}[t]{0.31\linewidth}
    \centering
    \includegraphics[width=0.98\textwidth]{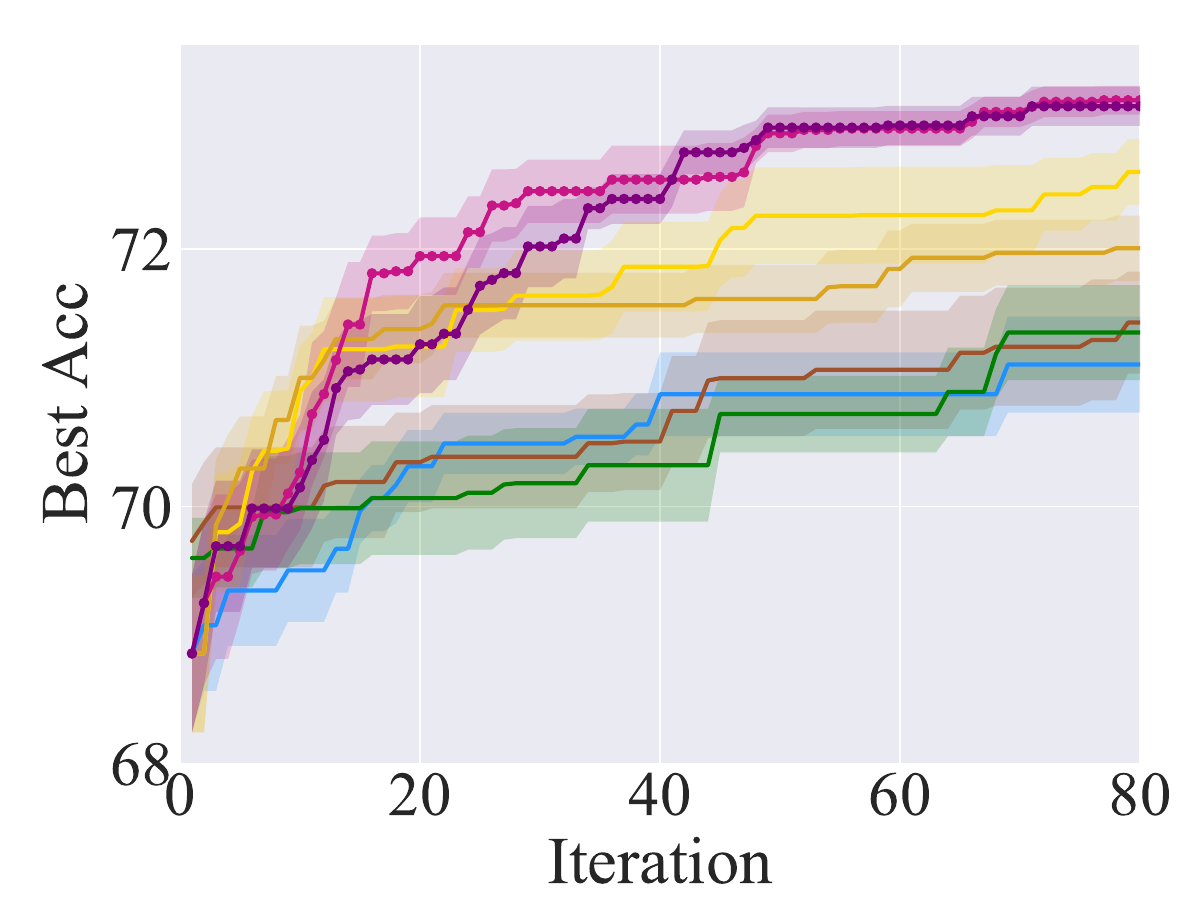}  
    \caption{CIFAR-100 valid-set}
    \label{supp:fig:cifar100_valid}
  \end{subfigure}
  \begin{subfigure}[t]{0.31\linewidth}
    \centering
    \includegraphics[width=0.98\textwidth]{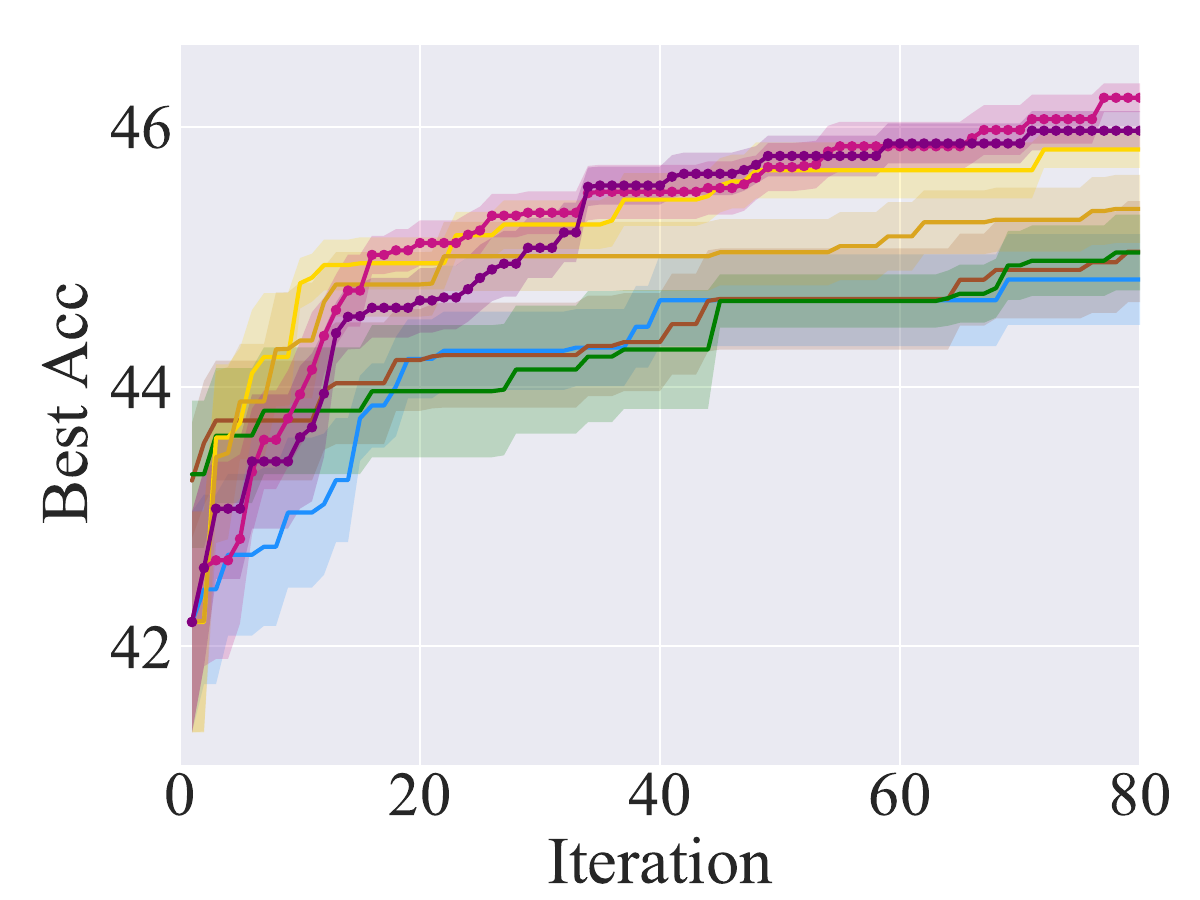}  
    \caption{ImageNet-16-120 valid-set}
    \label{supp:fig:imagenet_valid}
  \end{subfigure}
  \begin{subfigure}[t]{0.31\linewidth}
    \centering
    \includegraphics[width=0.98\textwidth]{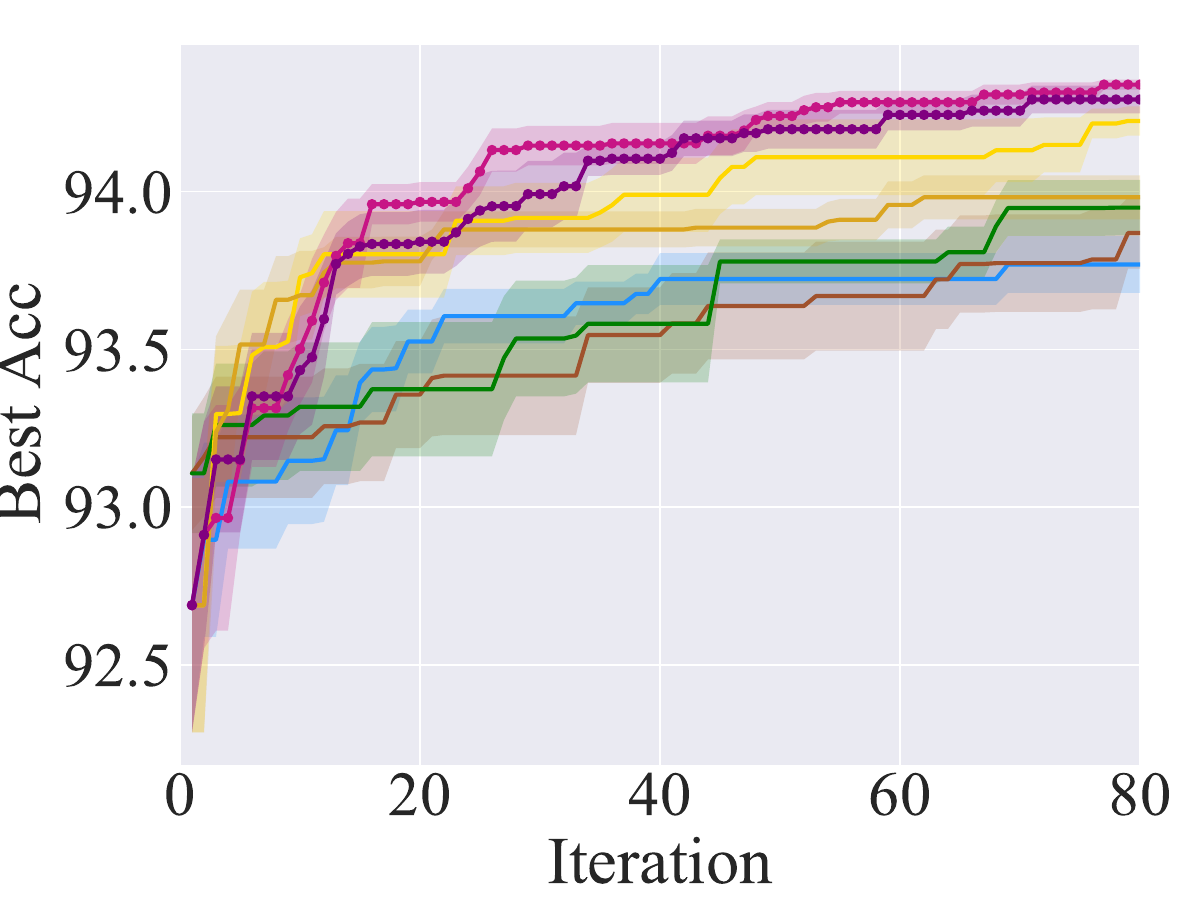}  
    \caption{CIFAR-10 test-set}
    \label{supp:fig:cifar10_test}
  \end{subfigure}
  \begin{subfigure}[t]{0.31\linewidth}
    \centering
    \includegraphics[width=0.98\textwidth]{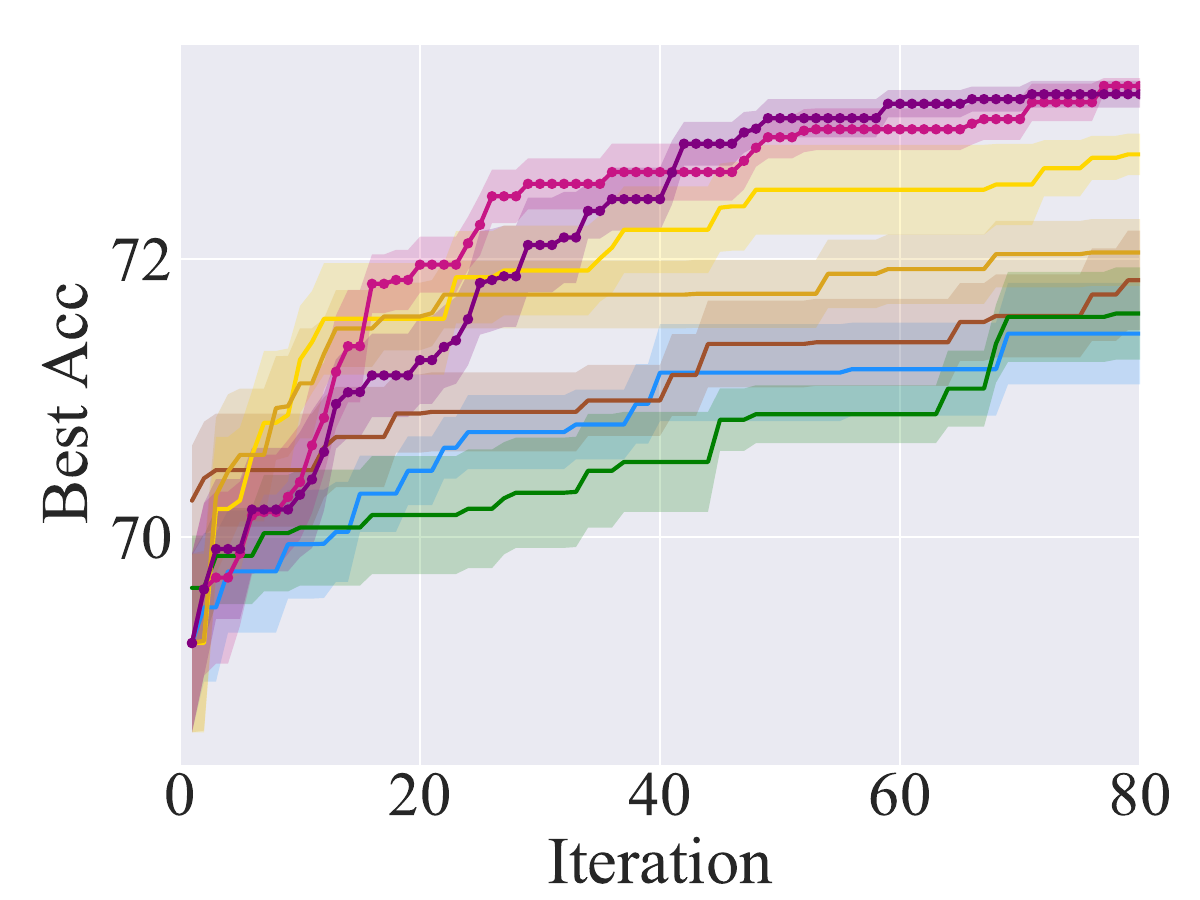}  
    \caption{CIFAR-100 test-set}
    \label{supp:fig:cifar100_test}
  \end{subfigure}
  \begin{subfigure}[t]{0.31\linewidth}
    \centering
    \includegraphics[width=0.98\textwidth]{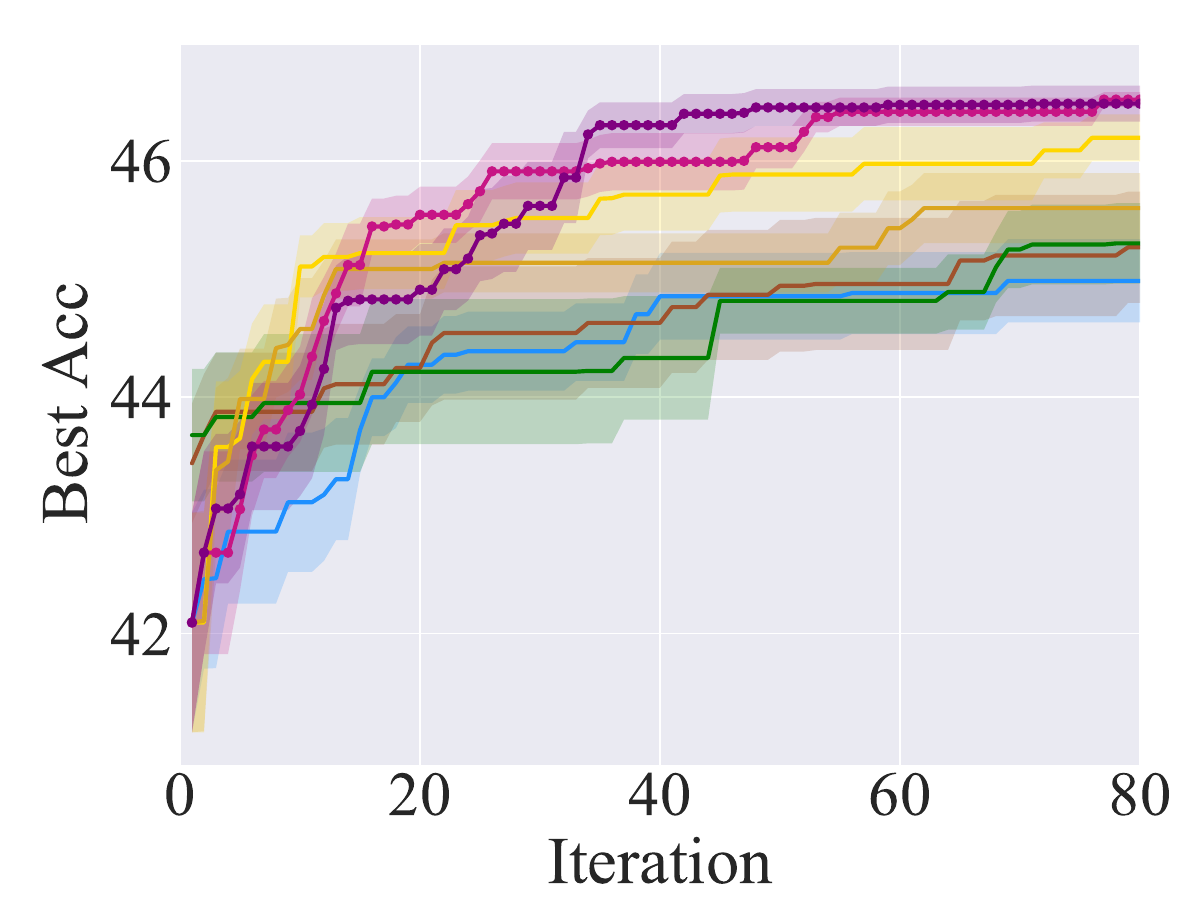}  
    \caption{ImageNet-16-120 test-set}
    \label{supp:fig:imagenet_test}
  \end{subfigure}
\vspace{-12pt}
\caption{Performance trend of random search and Bayesian optimization methods on NAS-Bench-201.
We run 10 times for each setting and plot the mean accuracy as the lines. Note that when testing Random Search, GP-BO, and our PoPBO, we adopt the same initial random seeds for all settings at each run for fairness. Hence, the lines in each plot have the same initial point (at the 0-th iteration).}
\vspace{-5pt}
\label{supp:fig:nas-bench-201}
\end{figure*}

\begin{table*}[tb]
\centering
\resizebox{.99\textwidth}{!}{
\smallskip\begin{tabular}{lcccc}
\toprule
\textbf{Methods} & \textbf{Naval} & \textbf{Parkinson} & \textbf{Protein} & \textbf{Slice} \\
\midrule
Random Search~\citep{DBLP:rs_journals/jmlr/BergstraB12}  &
$9.54 \times 10^{-5}\pm 5.76 \times 10^{-5}$ &
$7.99 \times 10^{-3}\pm 4.18 \times 10^{-3}$ &
$3.04 \times 10^{-2}\pm 1.51 \times 10^{-2}$ &
$1.48 \times 10^{-4}\pm 7.45 \times 10^{-5}$ \\
TPE~\citep{DBLP:tpe_conf/nips/BergstraBBK11}  &
$1.67 \times 10^{-5} \pm 1.64 \times 10^{-5}$ &
$4.49 \times 10^{-3} \pm 2.55 \times 10^{-3}$ &
$1.86 \times 10^{-2} \pm 1.37 \times 10^{-2}$ &
$1.80 \times 10^{-4} \pm 7.10 \times 10^{-5}$ \\
SMAC~\citep{DBLP:smac_conf/lion/HutterHL11}  &
$2.94 \times 10^{-4} \pm 3.96 \times 10^{-4}$ &
$1.48 \times 10^{-2} \pm 9.38 \times 10^{-3}$ &
$2.05 \times 10^{-2} \pm 1.14 \times 10^{-2}$ &
$3.16 \times 10^{-4} \pm 2.25 \times 10^{-4}$ \\
GP (EI)~\citep{DBLP:spearmint_conf/nips/SnoekLA12}  &
$1.79 \times 10^{-4} \pm 1.77 \times 10^{-4}$ &
$6.20 \times 10^{-3} \pm 2.44 \times 10^{-3}$ &
$8.06 \times 10^{-3} \pm 6.53 \times 10^{-2}$ &
$1.66 \times 10^{-4} \pm 7.00 \times 10^{-5}$ \\
GP (LCB)~\citep{DBLP:spearmint_conf/nips/SnoekLA12}  &
$1.38 \times 10^{-4} \pm 1.77 \times 10^{-4}$ &
$5.31 \times 10^{-3} \pm 2.79 \times 10^{-3}$ &
$8.08 \times 10^{-3} \pm 1.02 \times 10^{-2}$ &
$1.95 \times 10^{-4} \pm 1.30 \times 10^{-4}$ \\
\midrule
PoPBO (ERI)  &
$1.38 \times 10^{-5} \pm 1.59 \times 10^{-5}$ &
\bm{$1.92 \times 10^{-3} \pm 1.51 \times 10^{-3}$} &
$5.77 \times 10^{-3} \pm \bm{3.55 \times 10^{-3}}$ &
$4.83 \times 10^{-5} \pm 3.29 \times 10^{-5}$ \\
PoPBO (R-LCB)  &
\bm{$1.07 \times 10^{-5} \pm 6.26 \times 10^{-6}$} &
$2.41 \times 10^{-3} \pm 1.93 \times 10^{-3}$ &
$\bm{4.62 \times 10^{-3}} \pm 3.91 \times 10^{-3}$ &
\bm{$3.14 \times 10^{-5} \pm 1.74 \times 10^{-5}$} \\
\bottomrule
\end{tabular}
}
\vspace{-5pt}
\caption{Regret of the configuration discovered by various methods on the four datasets of HPO-Bench. We run each method for ten times and report the mean and standard deviation. The best performance (lowest mean value and standard deviation) is in bold.}
\label{supp:tab:hpo_bench}
\vspace{-8pt}
\end{table*}
\section{Supplementary of Experiments}
\subsection{Experimental Settings on Robustness Analysis}~\label{supp:sub:forrester}
We compare the sensitivity to additive Gaussian noise between GP (value-based) response surface and PoPBO (ranking-based) response surface on Forrester function.
To simulate the performance of GP, we first utilize Gaussian process to fit a certain number of (15 in this paper) observed values and plot the ranking of e.g. 100 points according to the values predicted by GP as Fig.~\ref{fig:compare_gp}. Meanwhile, our method utilizes Poisson process to directly capture the ranking response surface based on the same 15 observations and predict the ranking of 100 points as shown in Fig.~\ref{fig:compare_pp}. 
We observe that our response surface is more robust to noise and can better capture the global ranking.

\subsection{Details of the Benchmarks}\label{supp:sec:benchmark}
For the simulated benchmark, we apply PoPBO to optimize three simulation functions: 1) 2-d Branin function with the domains of each dimension are $[-5, 10]$ and $[0, 15]$   respectively; 2) 6-d Hartmann function in $[0, 1]$ for all six dimensions; 3) 6-d Rosenbrock function defined in $[-5, 10]^6$.
For the HPO task, we test PoPBO on the tabular benchmark HPO-Bench~\citep{DBLP:hpobench_conf/nips/EggenspergerMMF21}, containing the root mean square error (RMSE) of a 2-layer feed-forward neural network (FCNET)~\citep{DBLP:fcnet_journals/corr/abs-1905-04970} trained under 62208 hyper-parameter configurations on four real-world datasets: protein structure~\citep{protein_rana2013physicochemical}, slice localization~\citep{DBLP:slice_conf/miccai/GrafKSPC11}, naval propulsion~\citep{naval_coraddu2016machine} and parkinsons telemonitoring~\citep{DBLP:parkinson_journals/tbe/TsanasLMR10}.
The averaged RMSE over four independent runs under the same configuration is utilized as the performance of that configuration.
For the NAS task, we test on NAS-Bench-201~\citep{DBLP:nasbench201_conf/iclr/Dong020} containing 15,625 architectures in a cell search space that consists of 6 categorical parameters, and each parameter has five choices. Each architecture is evaluated on three datasets. Following the setting of~\citep{DBLP:nasbench201_conf/iclr/Dong020}, we search for the best architecture according to its performance on the CIFAR-10 validation set after 12 epochs training.

\subsection{Detailed Settings of Baseline Methods}~\label{supp:subsec:baselines}
\vspace{-5pt}
In this section, we provide the specific details of each baseline mentioned in the paper:

\paragraph{Random Search (RS)} Following the description in~\cite{DBLP:rs_journals/jmlr/BergstraB12}, we sample candidates uniformly at random.

\paragraph{BO with Gaussian Process (GP)} We follow the settings that described by~\cite{DBLP:spearmint_conf/nips/SnoekLA12} and use the implementation of our own. We use expected improvement (EI) and lower confidence bound (LCB) as acquisition functions and adopt LBFGS to optimize them. When the search space is completely discrete like~\cite{DBLP:nasbench201_conf/iclr/Dong020}, we use random sampling to find the next query, which gives the maximizer of the acquisition function among $N=1000$ random samples. For kernel function, we use Mat\'ern 5/2 kernel for GP. During the training process, we adopt slice sampling, an efficient Markov chain Monte Carlo (MCMC) method, to fit the hyperparameters of GP, which we find to work more robustly for GP. 

\paragraph{Tree Parzen Estimator (TPE)}~\cite{DBLP:tpe_conf/nips/BergstraBBK11} adopt kernel density estimators to model the probability of points with bad and good performance respectively. Then TPE give the next query by optimizing the ratio between the two estimated likelihood, which is proved to be equivalent to optimizing EI. We use the default settings provided in hyperopt package (\url{https://github.com/hyperopt/hyperopt}).

\paragraph{SMAC}~\cite{DBLP:smac_conf/lion/HutterHL11} adopt random forest to model the response surface of the black-box function. We use the default settings given by scikit-optimize package (\url{https://github.com/scikit-optimize/scikit-optimize}).

\paragraph{BOHAMIANN} Unlike~\cite{DBLP:spearmint_conf/nips/SnoekLA12}, BOHAMIANN adopts Bayesian neural network to build the response surface, whose weights are sampled via a stochastic gradient Hamiltonian Monte-Carlo (SGHMC) method. We use the default settings provided in pybnn package (\url{https://github.com/automl/pybnn}) and use EI as the acquisition function.

\paragraph{PPBO}
This is an effective preferential BO method based on pairwise comparisons, attempting to learn user preferences in high-dimensional spaces. We use the default settings provided by PPBO package (\url{https://github.com/AaltoPML/PPBO}).

\paragraph{HEBO} Heteroscedastic Evolutionary Bayesian Optimisation that
won the NeurIPS 2020 black-box optimisation competition. We use the default strategy and its default parameters provided in HEBO package (\url{https://github.com/huawei-noah/HEBO}). For a fair comparison, we use a uniform sampling strategy instead of a sobol one during initialization and candidates generation.

\subsection{Detailed Results on HPO-Bench and NAS-Bench-201} \label{supp:subsec:exp}

Table~\ref{supp:tab:hpo_bench} reports the numerical performance of PoPBO and other methods on the four datasets on HPO-Bench.
Fig.~\ref{supp:fig:nas-bench-201} displays the performance trend of PoPBO and other methods on the validation set and test set under the search space of NAS-Bench-201.

\begin{figure}
\centering
    \centering
  \begin{subfigure}[t]{0.48\linewidth}
    \centering
    \includegraphics[width=0.98\textwidth]{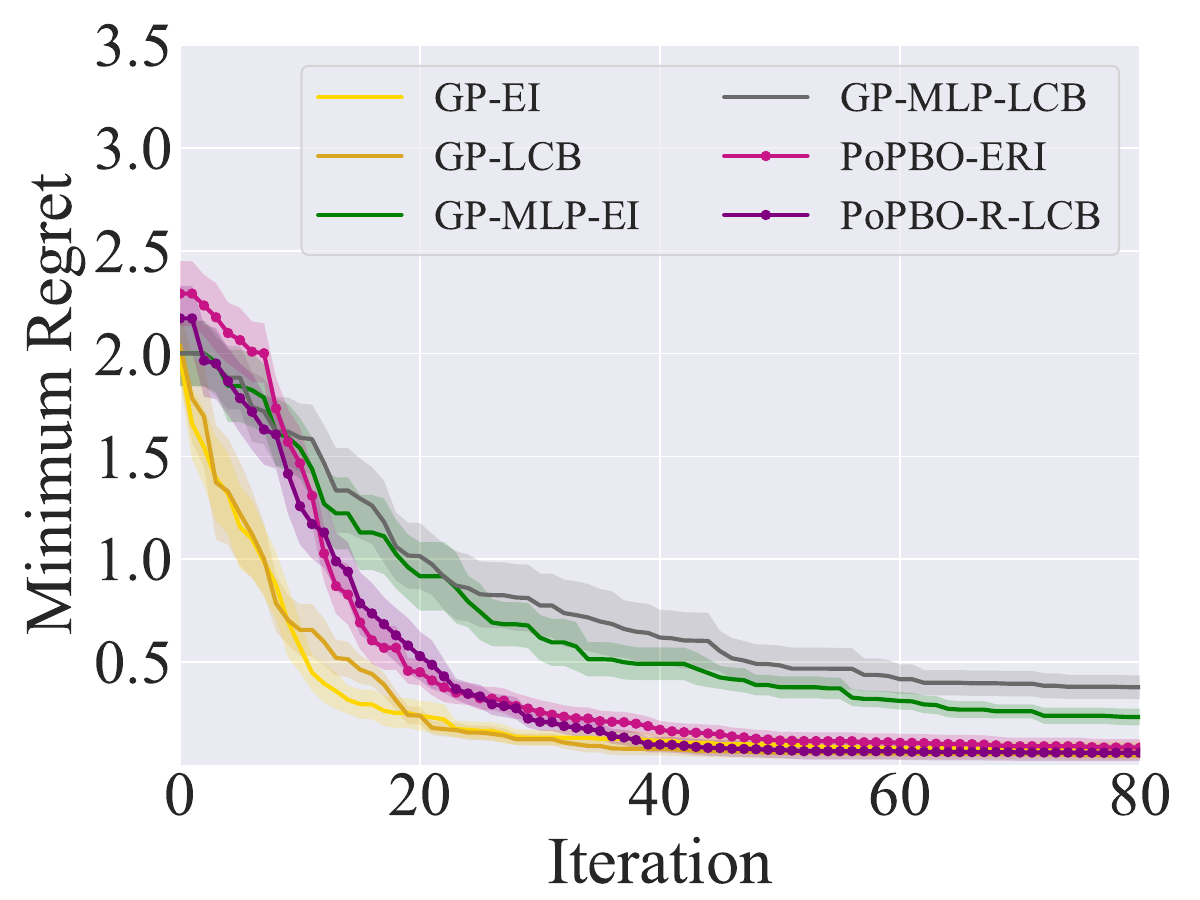}  
    \caption{6-d Hartmann}
    \label{fig:gp_mlp_hartmann}
  \end{subfigure}
  \begin{subfigure}[t]{0.48\linewidth}
    \centering
    \includegraphics[width=0.98\textwidth]{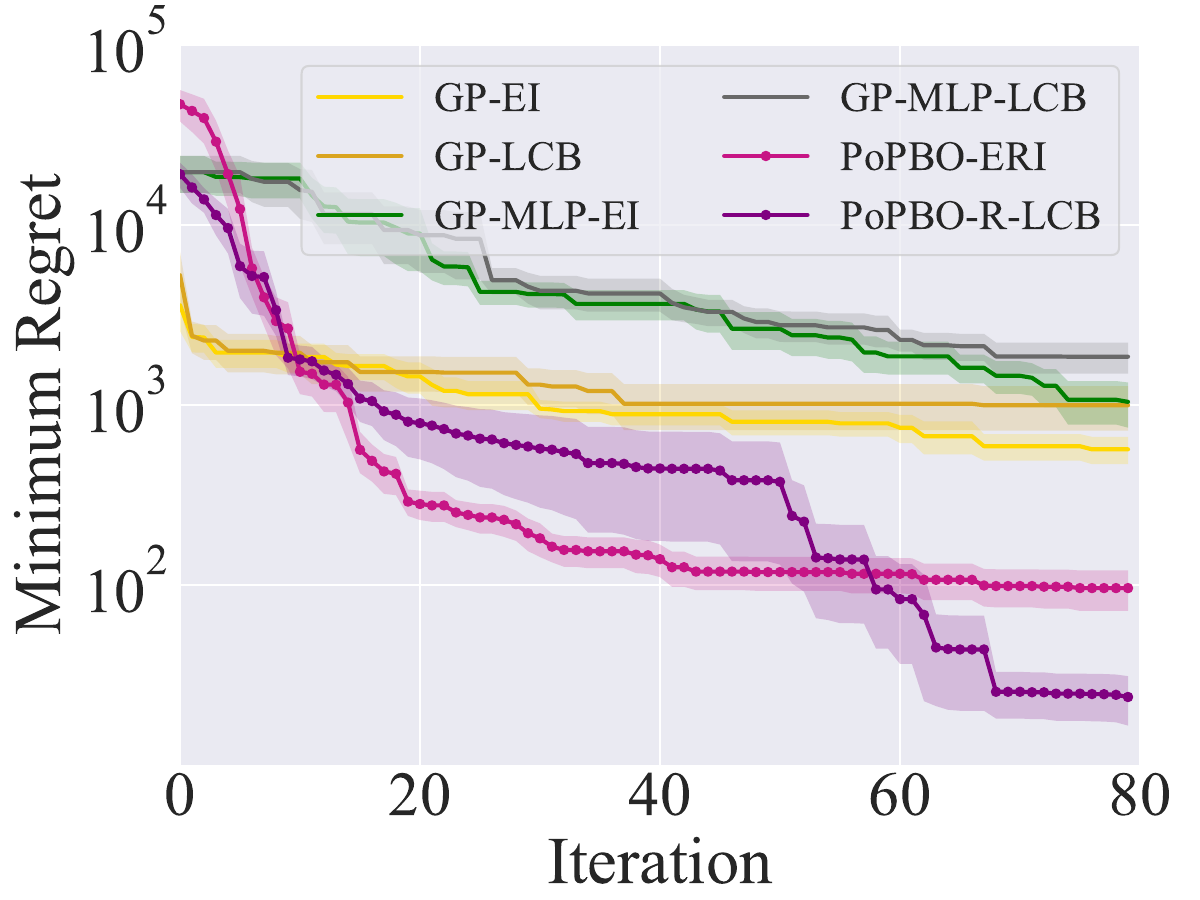}  
    \caption{6-d Rosenbrock}
    \label{fig:gp_mlp_rosenbrock}
  \end{subfigure}
\vspace{-12pt}
\caption{Performance trend of GP, GP-MLP, and PoPBO on Hartmann and Rosenbrock simulated benchmarks.}
\label{fig:gp_mlp}
\vspace{-10pt}
\end{figure}

\subsection{Is the gain in performance of PoPBO due to the complex representation of MLP in the surrogate model?}
We utilize the same MLP architecture and training settings as PoPBO's to fit the Gaussian likelihood of each candidate. We denote such a setting as GP-MLP.
We plot the regret of GP-MLP, GP, and our PoPBO on Hartmann and Rosenbrock in Fig.~\ref{fig:gp_mlp}. We observe that GP-MLP performs much worse than GP and ours, showing that the complex representation of the MLP in the surrogate model is not the main reason for the gain in performance.

\subsection{Ablation Study about the Worst Tolerant Ranking $K_m$ in ERI}

\begin{figure*}[tb]
    \centering
  \begin{subfigure}[t]{0.48\linewidth}
    \centering
    \includegraphics[width=0.98\textwidth]{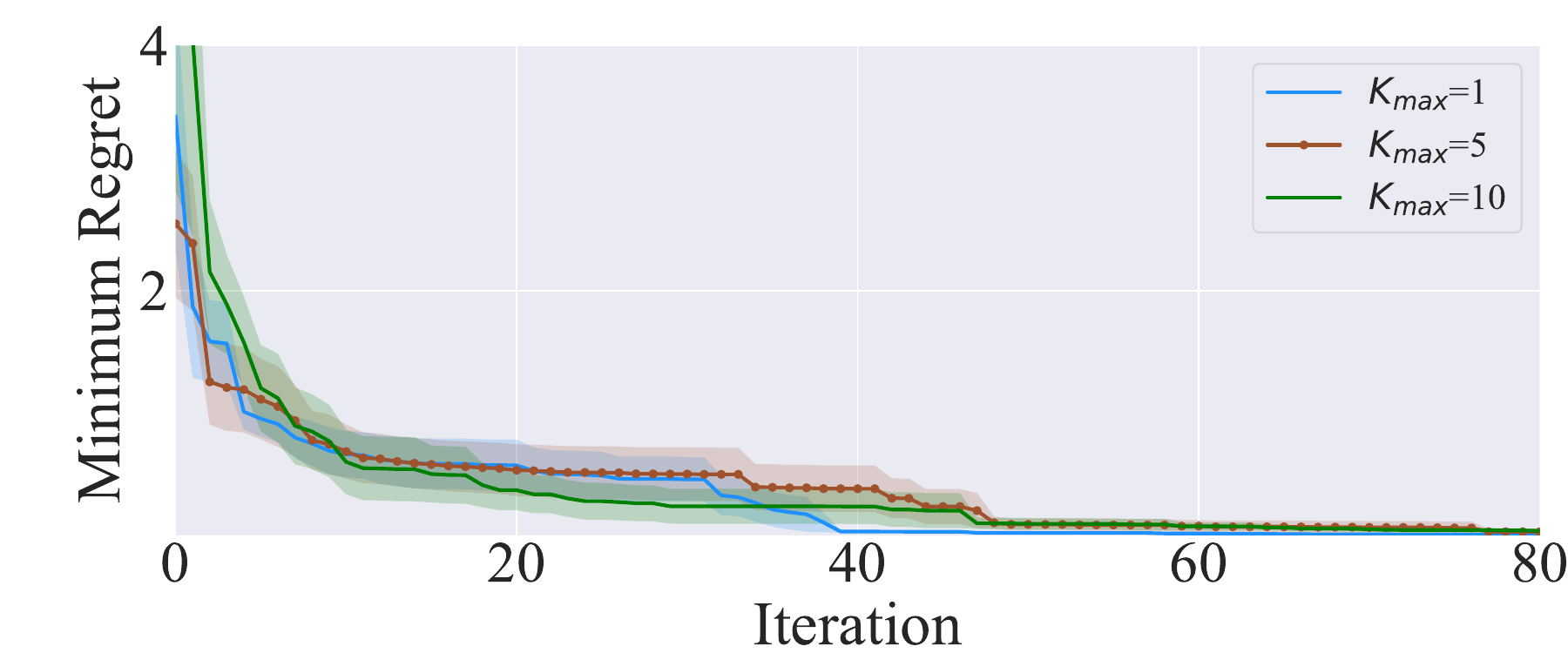}  
    \caption{2-d Branin}
    \label{fig:ab_eri_branin_k}
  \end{subfigure}
  \begin{subfigure}[t]{0.48\linewidth}
    \centering
    \includegraphics[width=0.98\textwidth]{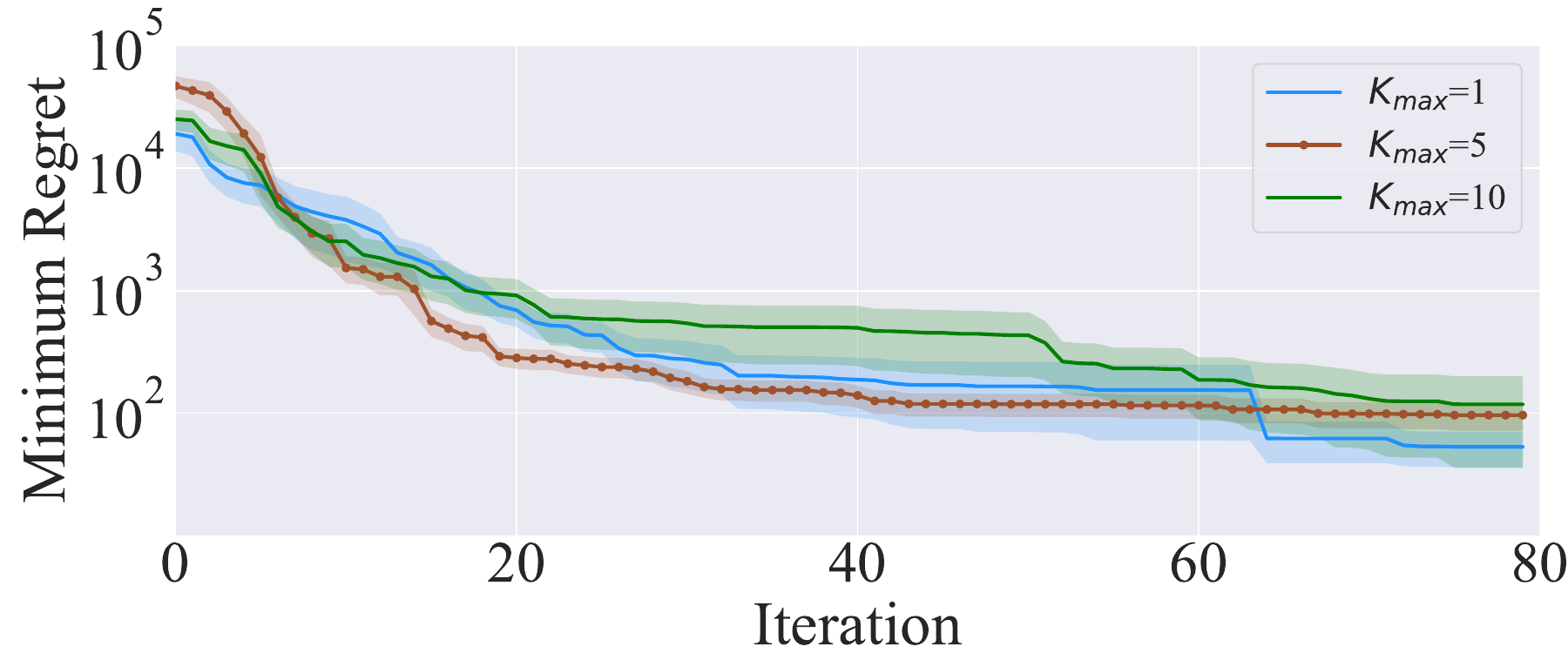}  
    \caption{6-d Rosenbrock}
    \label{fig:ab_eri_rosenbrock_k}
  \end{subfigure}
\vspace{-12pt}
\caption{Ablation study of ERI on the two simulation functions. $K_{max}$ is the worst tolerable ranking, and a higher $K_{max}$ leads to a higher rate of exploration.}
\vspace{-5pt}
\label{fig:ablation_eri}
\end{figure*}

Fig.~\ref{fig:ablation_eri} compares the performance of ERI under various settings for the worst tolerant ranking $K_{m}$. 
We observe that PoPBO-ERI is not very sensitive to $K_m$.
Specifically, for Branin, ERI with larger $K_m$ converges faster but ERI with various $K_m$ have similar ultimate performance after 80 iterations. 
For Rosenbrock that has larger search space and more complex landscape, better exploitation ability is more important than exploration, and thus ERI with lower $K_m$ achieves better performance.
\subsection{Ablation Study about the Rectified Technique on NAS-Bench-201} \label{supp:sub:nasbench_ab}

Table~\ref{tab:nasbench201_ablation} evaluates the effect of quantile parameters $q$ on both R-LCB and ERI on NAS-Bench-201.  We adopt the same ten random seeds for different settings for fair comparison. The results show that our settings of q (0.4 for ERI and 0.6 for R-LCB) perform still well on the real-world benchmark. Moreover, we observe that the results are the same when $q=0.4, 0.5, 0.6$. We analyze that ERI and R-LCB have the same effect as the vanilla EI and LCB respectively when $q\geq0.4$. 

\begin{table*}[tb!]
\centering
\caption{Ablation study about rectified hyperparameter $q$ on NAS-Bench-201.}
\vspace{-8pt}
\setlength{\tabcolsep}{2pt}
\resizebox{.85\textwidth}{!}{
\smallskip\begin{tabular}{lcccccc}
\toprule
\multirow{2}{*}{\textbf{Methods}} & \multicolumn{2}{c}{\textbf{CIFAR-10}}&  \multicolumn{2}{c}{\textbf{CIFAR-100}} & \multicolumn{2}{c}{\textbf{ImageNet-16-120}}  \\
 \cmidrule(lr){2-3}
 \cmidrule(lr){4-5}
 \cmidrule(lr){6-7}
 & \textbf{valid} & \textbf{test} &  \textbf{valid} & \textbf{test} & \textbf{valid} & \textbf{test} \\
\midrule
PoPBO (ERI, q=0.1)  &
91.41$\pm$0.15 &
94.22$\pm$0.15 &
72.70$\pm$0.63 &
72.68$\pm$0.69 &
45.93$\pm$0.61 &
46.45$\pm$0.34 \\
PoPBO (ERI, q=0.2)  &
91.47$\pm$0.12 &
94.26$\pm$0.16 &
72.96$\pm$0.59 &
73.00$\pm$0.53 &
45.90$\pm$0.66 &
46.33$\pm$0.55 \\
PoPBO (ERI, q=0.4)  &
91.52$\pm$0.04 &
94.33$\pm$0.08 &
73.21$\pm$0.36 &
73.19$\pm$0.31 &
46.12$\pm$0.43 &
46.61$\pm$0.32 \\
PoPBO (ERI, q=0.5)  &
91.52$\pm$0.04 &
94.33$\pm$0.08 &
73.21$\pm$0.36 &
73.19$\pm$0.31 &
46.12$\pm$0.43 &
46.61$\pm$0.32 \\
PoPBO (ERI, q=0.6)  &
91.52$\pm$0.04 &
94.33$\pm$0.08 &
73.21$\pm$0.36 &
73.19$\pm$0.31 &
46.12$\pm$0.43 &
46.61$\pm$0.32 \\

PoPBO (R-LCB, q=0.1)  &
91.33$\pm$0.19 &
94.19$\pm$0.23 &
72.62$\pm$0.82 &
72.52$\pm$0.67 &
45.71$\pm$0.56 &
46.10$\pm$0.65 \\
PoPBO (R-LCB, q=0.2)  &
91.50$\pm$0.06 &
94.30$\pm$0.11 &
73.04$\pm$0.56 &
73.04$\pm$0.45 &
46.07$\pm$0.38 &
46.37$\pm$0.30 \\
PoPBO (R-LCB, q=0.4)  &
91.52$\pm$0.04 &
94.33$\pm$0.08 &
73.21$\pm$0.36 &
73.19$\pm$0.31 &
46.12$\pm$0.43 &
46.61$\pm$0.32 \\
PoPBO (R-LCB, q=0.5)  &
91.52$\pm$0.04 &
94.33$\pm$0.08 &
73.21$\pm$0.36 &
73.19$\pm$0.31 &
46.12$\pm$0.43 &
46.61$\pm$0.32 \\
PoPBO (R-LCB, q=0.6)  &
91.52$\pm$0.04 &
94.33$\pm$0.08 &
73.21$\pm$0.36 &
73.19$\pm$0.31 &
46.12$\pm$0.43 &
46.61$\pm$0.32 \\

\bottomrule
\end{tabular}
}
\label{tab:nasbench201_ablation}
\vspace{-3pt}
\end{table*}

\begin{figure}
\centering
    \includegraphics[width=0.4\textwidth]{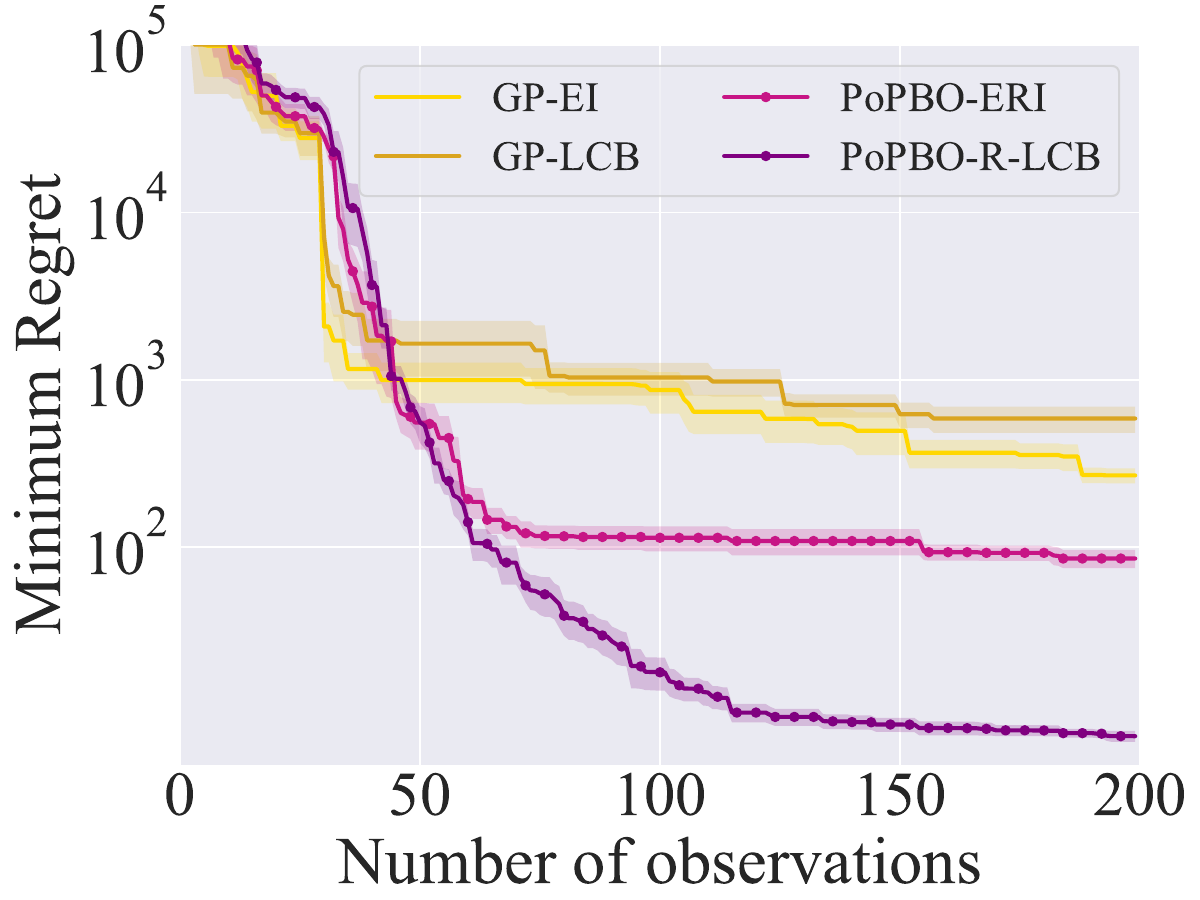} 
\vspace{-12pt}
\caption{Performance trend of PoPBO and GP by running 200 iterations on 6-d Rosenbrock. For each setting, we conduct replicated experiments for six times with various random seeds.}
\label{fig:rosenbrock_many}
\vspace{-10pt}
\end{figure}

\begin{figure}[htbp]
    \centering
    \includegraphics[width=0.6\textwidth]{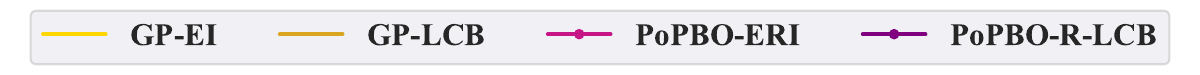}\vskip -5pt
  \begin{subfigure}[t]{0.31\linewidth}
    \centering
    \includegraphics[width=0.98\textwidth]{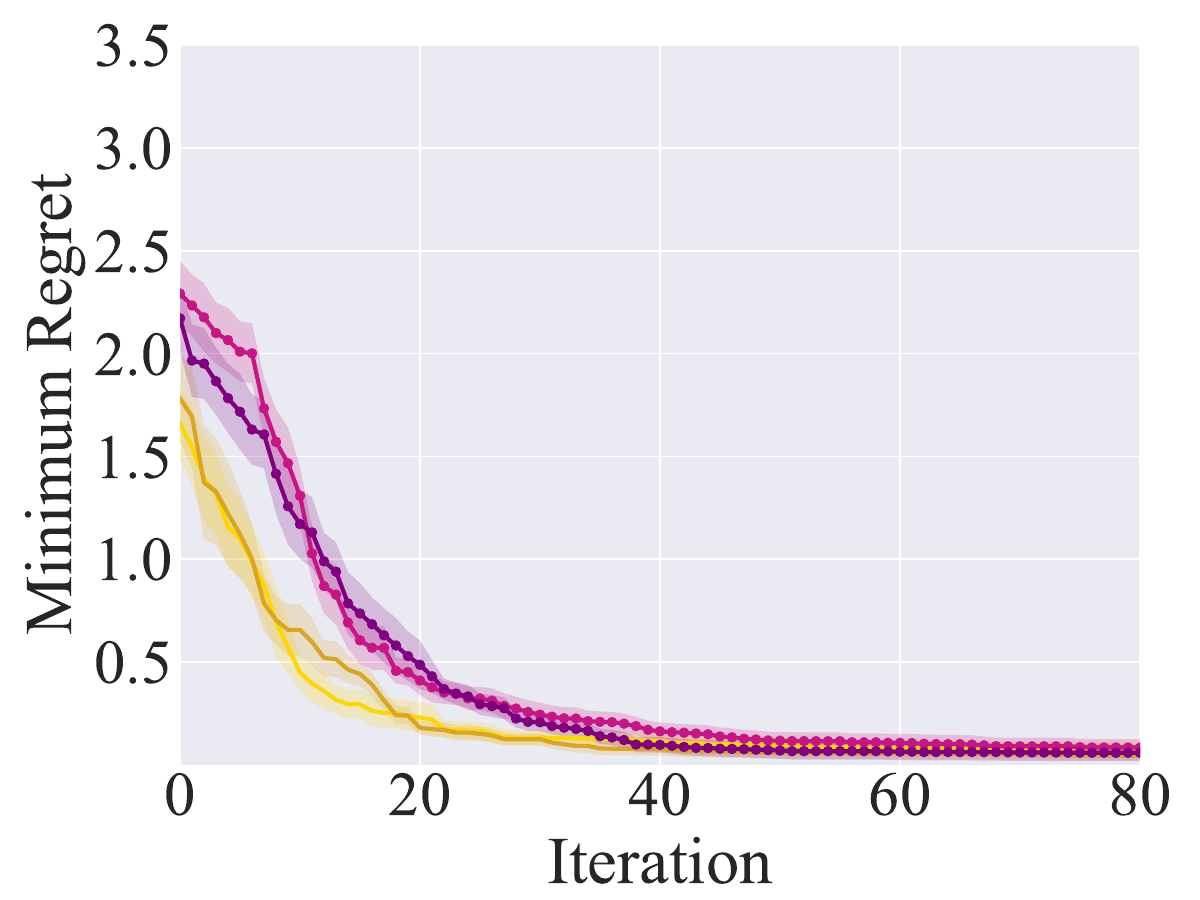}  
    \caption{6-d Hartmann ($\sigma=0$)}
    \label{fig:noise_hartmann_0.05}
  \end{subfigure}
  \begin{subfigure}[t]{0.31\linewidth}
    \centering
    \includegraphics[width=0.98\textwidth]{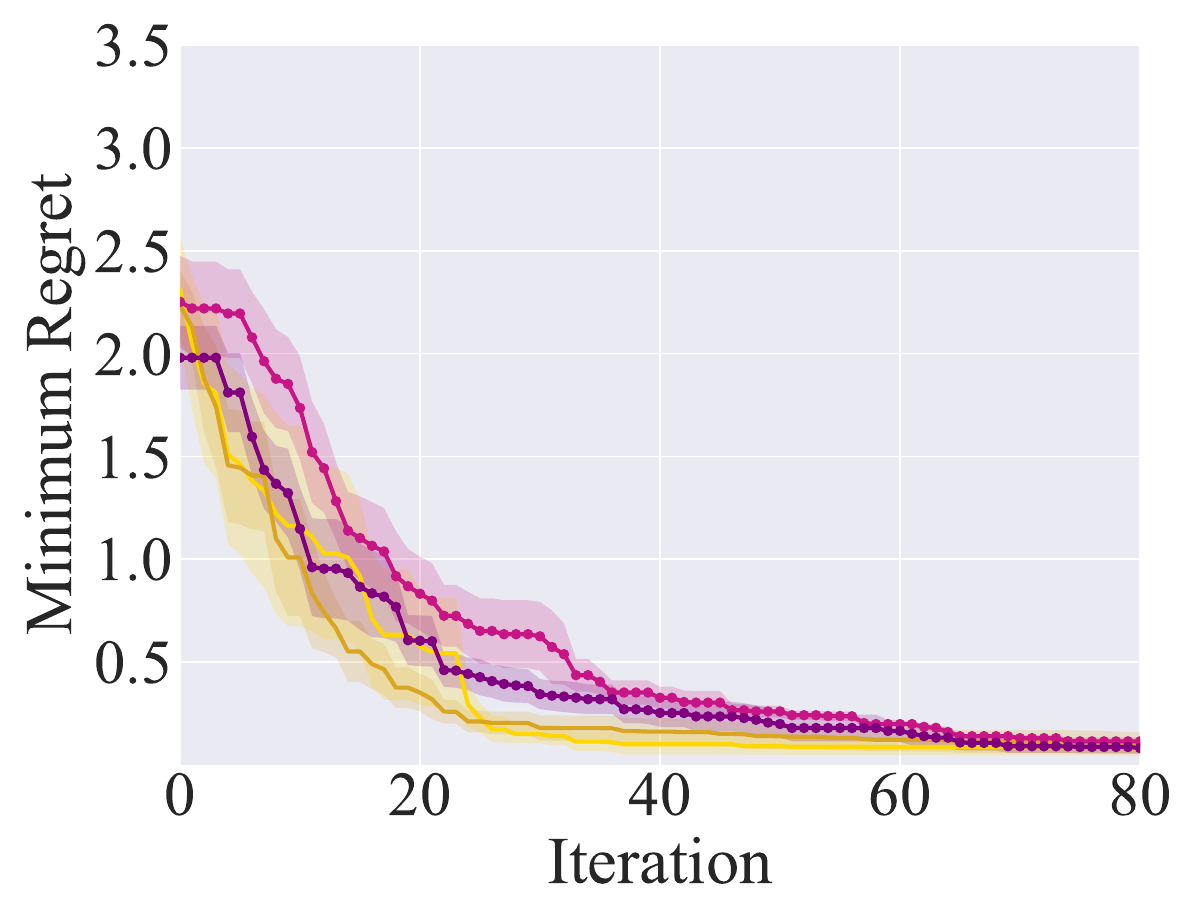}  
    \caption{6-d Hartmann ($\sigma=0.075$)}
    \label{fig:noise_hartmann_0.75}
  \end{subfigure}
  \begin{subfigure}[t]{0.31\linewidth}
    \centering
    \includegraphics[width=0.98\textwidth]{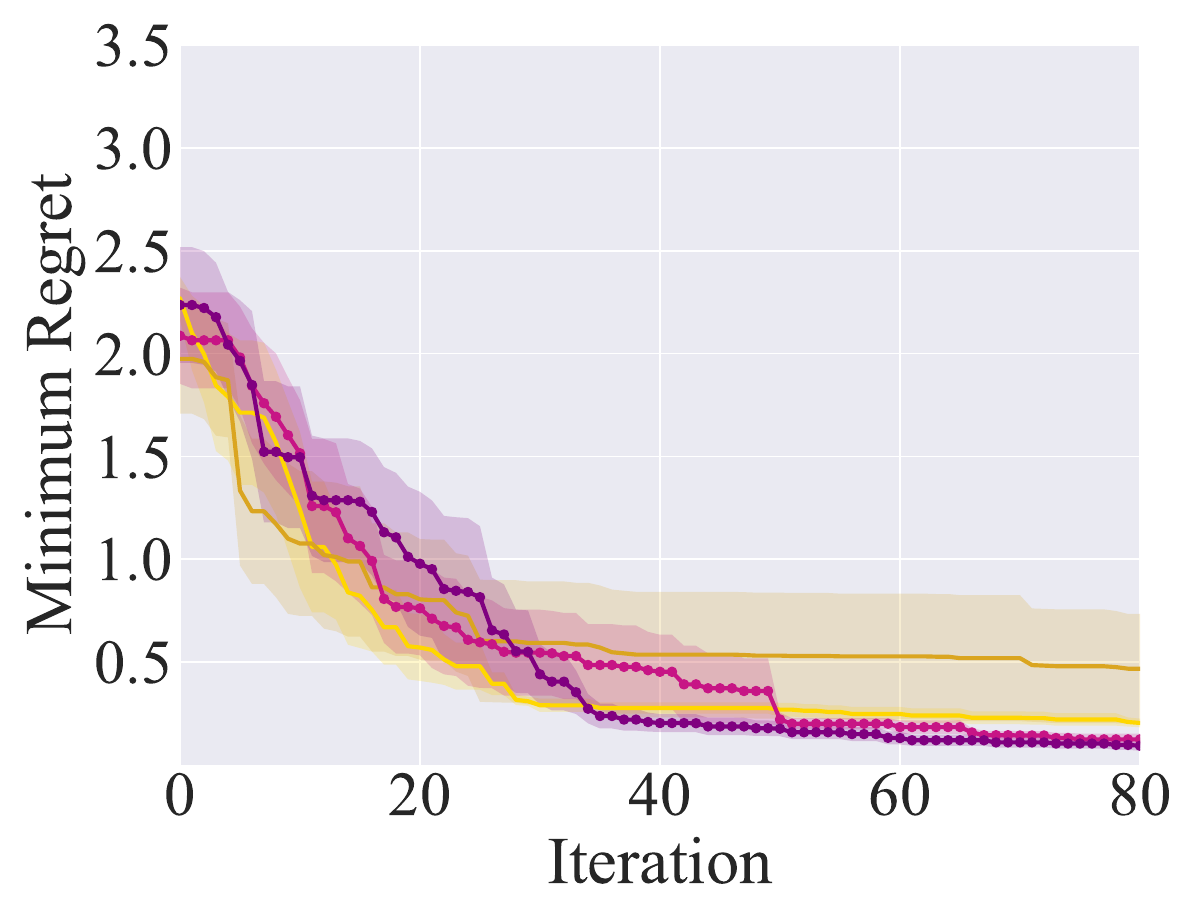}  
    \caption{6-d Hartmann ($\sigma=0.1$)}
    \label{fig:noise_hartmann_0.1}
  \end{subfigure}

  \begin{subfigure}[t]{0.31\linewidth}
    \centering
    \includegraphics[width=0.98\textwidth]{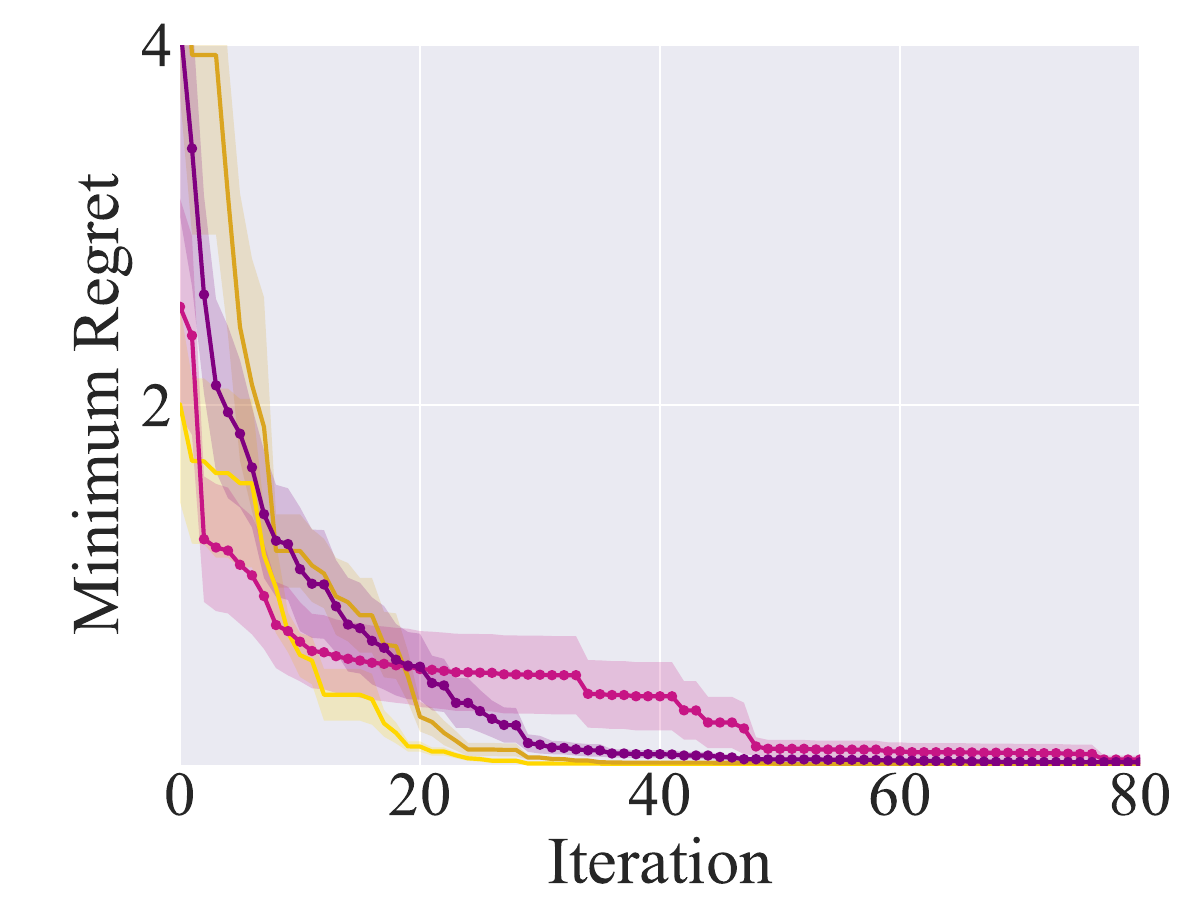}  
    \caption{2-d Branin ($\sigma=0$)}
    \label{fig:noise_branin_0}
  \end{subfigure}
  \begin{subfigure}[t]{0.31\linewidth}
    \centering
    \includegraphics[width=0.98\textwidth]{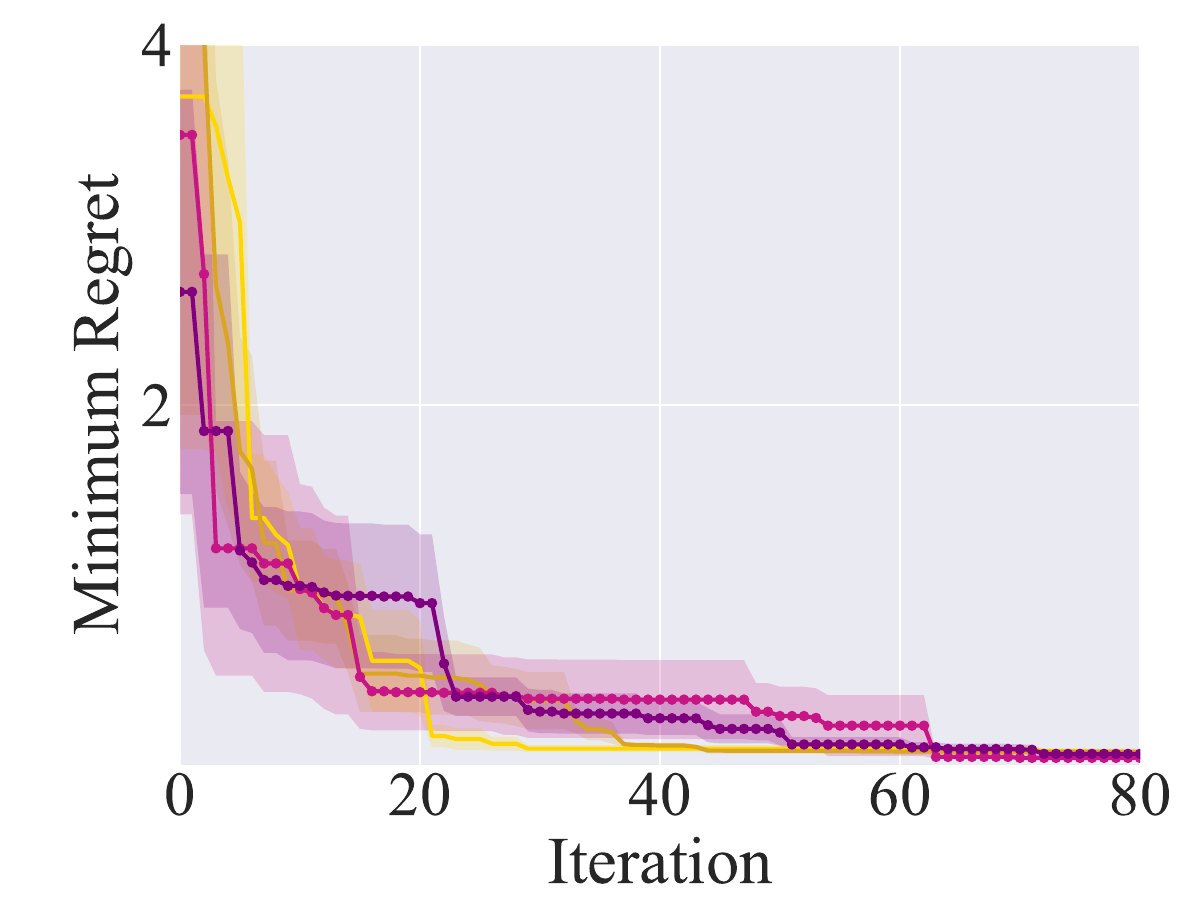}  
    \caption{2-d Branin ($\sigma=2$)}
    \label{fig:noise_branin_2}
  \end{subfigure}
  \begin{subfigure}[t]{0.31\linewidth}
    \centering
    \includegraphics[width=0.98\textwidth]{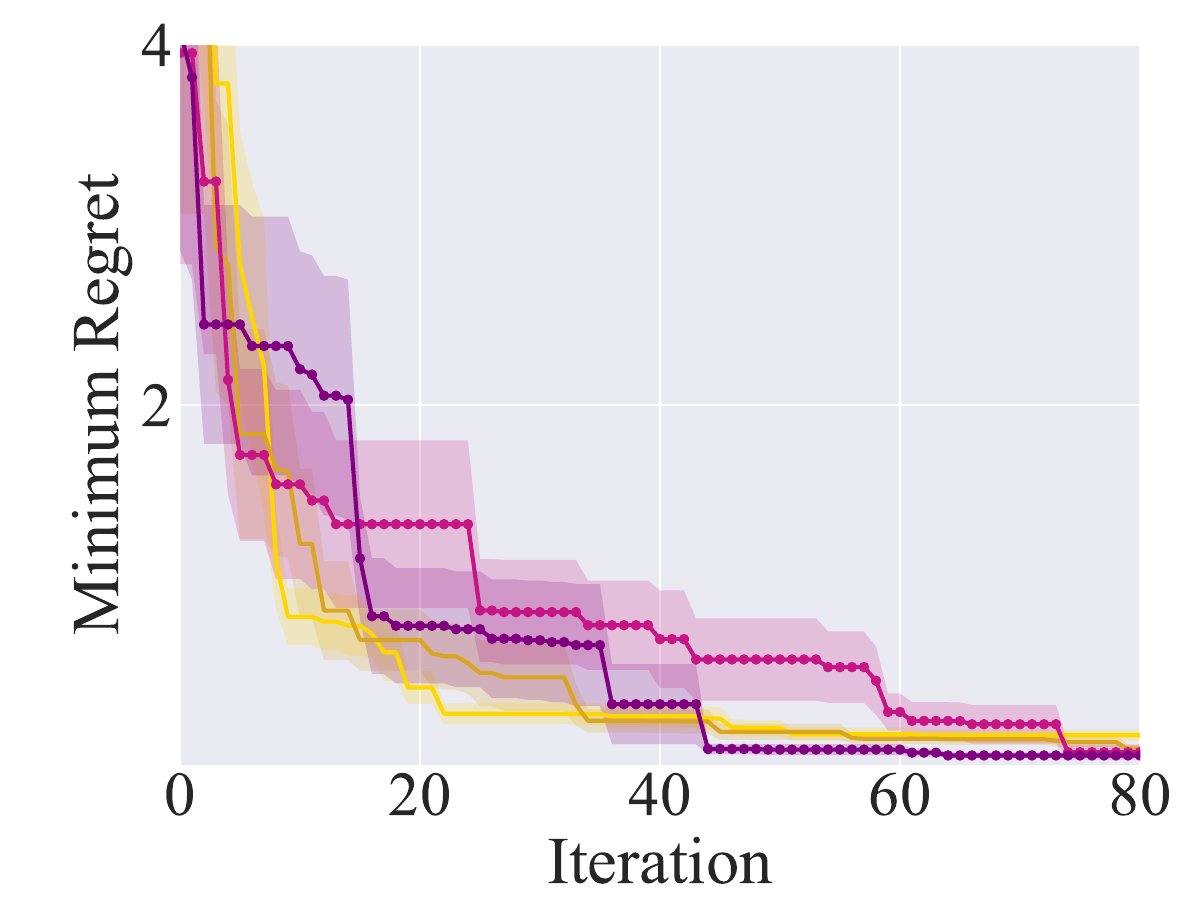}  
    \caption{2-d Branin ($\sigma=5$)}
    \label{fig:noise_branin_5}
  \end{subfigure}
  
  \begin{subfigure}[t]{0.31\linewidth}
    \centering
    \includegraphics[width=0.98\textwidth]{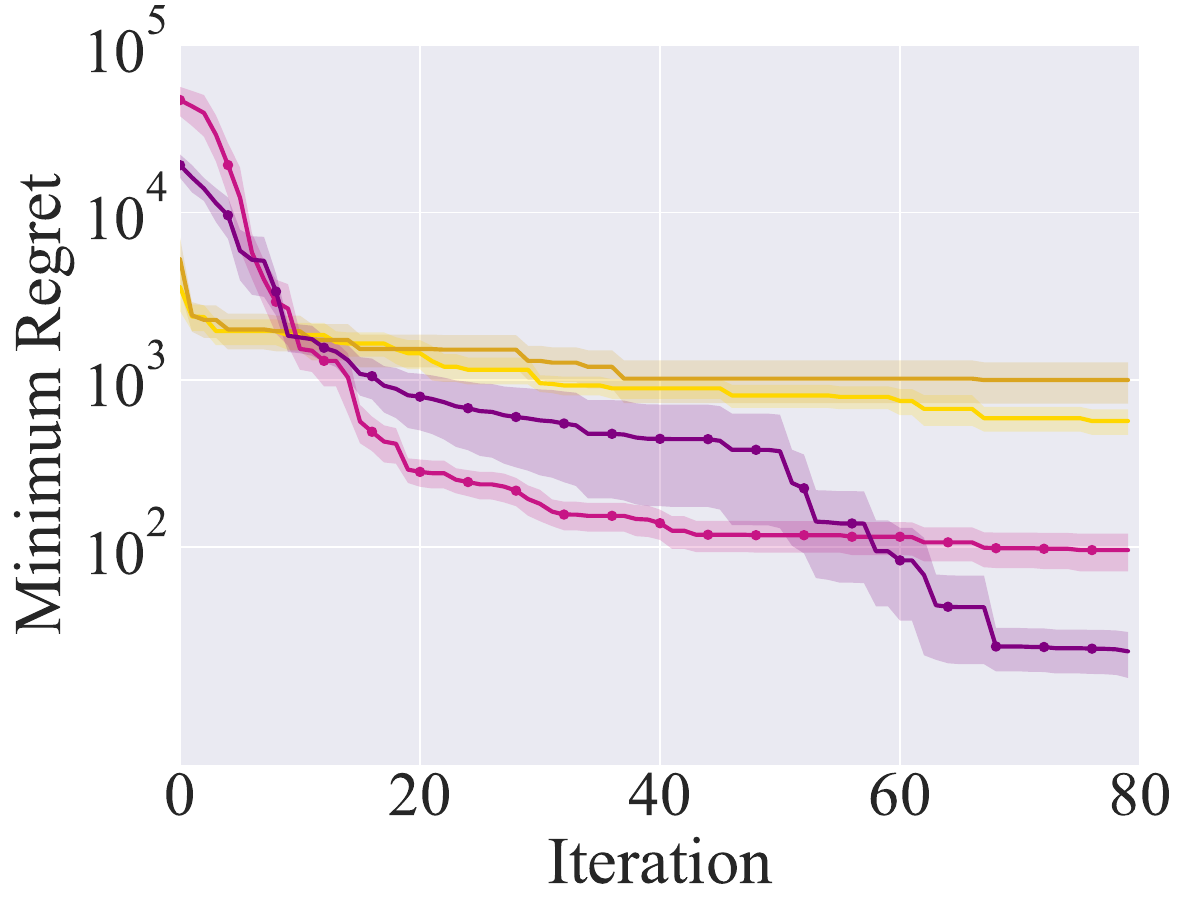}  
    \caption{2-d Rosenbrock ($\sigma=0$)}
    \label{fig:noise_rosenbrock_0}
  \end{subfigure}
  \begin{subfigure}[t]{0.31\linewidth}
    \centering
    \includegraphics[width=0.98\textwidth]{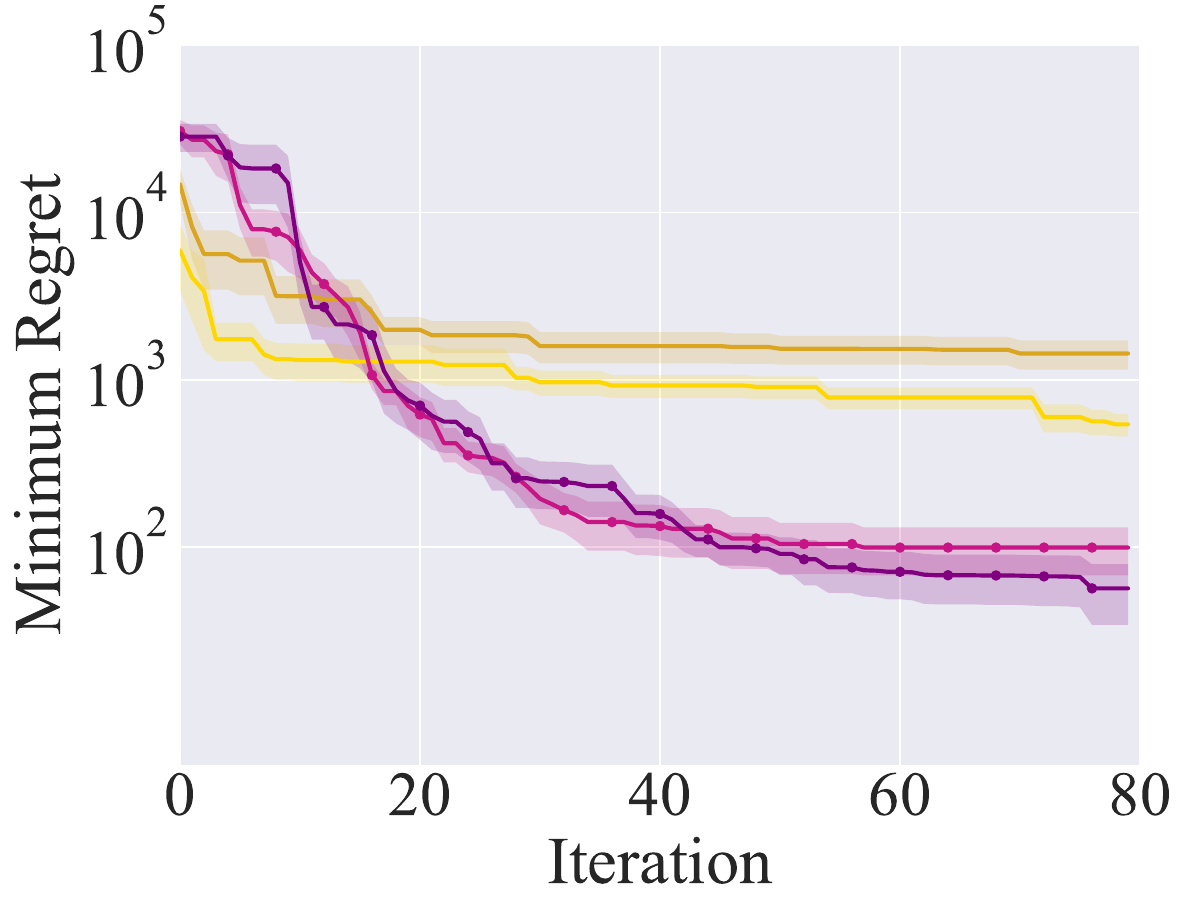}  
    \caption{2-d Rosenbrock ($\sigma=2$)}
    \label{fig:noise_rosenbrock_2}
  \end{subfigure}
  \begin{subfigure}[t]{0.31\linewidth}
    \centering
    \includegraphics[width=0.98\textwidth]{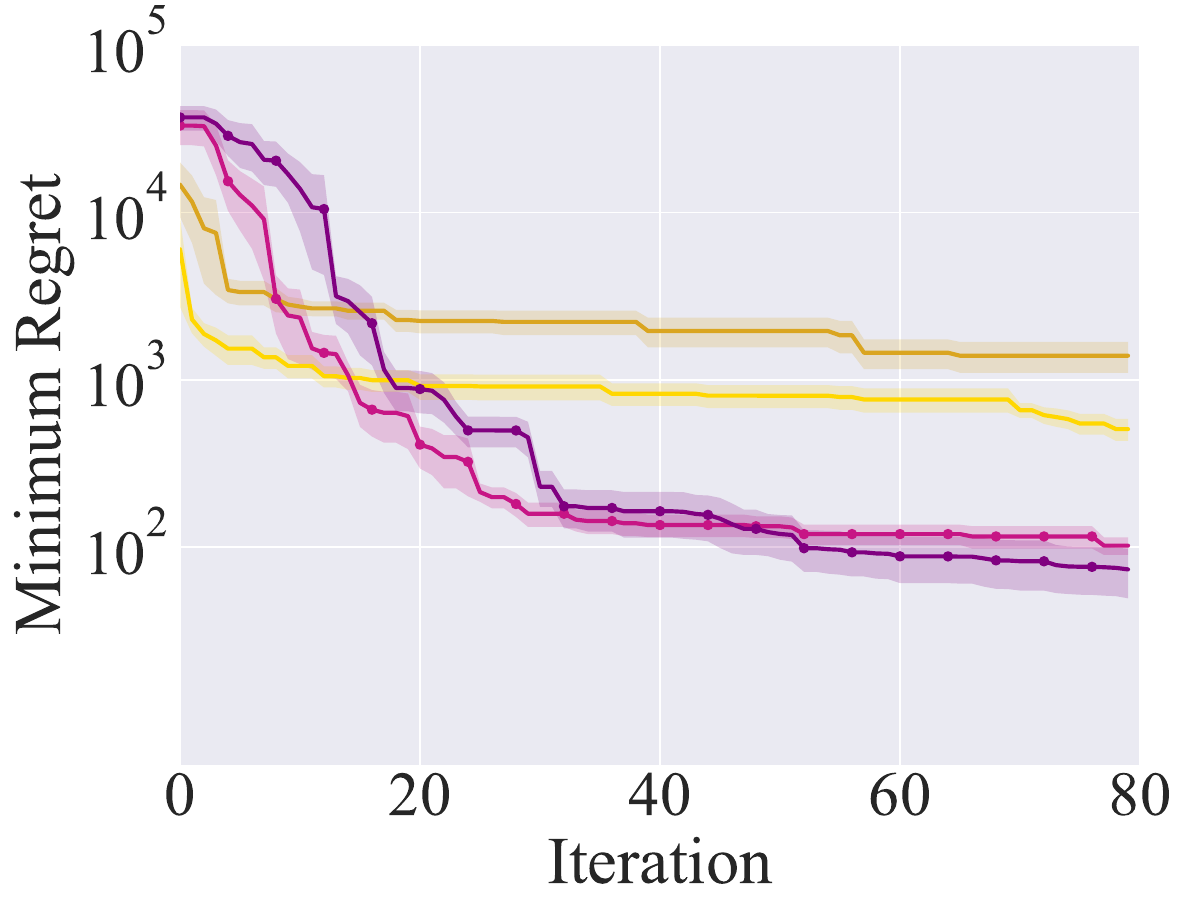}  
    \caption{2-d Rosenbrock ($\sigma=5$)}
    \label{fig:noise_rosenbrock_5}
  \end{subfigure}

\vspace{-10pt}
\caption{Comparison of the robustness to noise between GP-BO and PoPBO. We add Gaussian noises with zero mean and various standard deviation $\sigma$ to the objective function. Notice that Branin and Rosenbrock have much larger range of value than Hartman, so we add noises with larger variance to them.}
\vspace{-5pt}
\label{fig:noise}
\end{figure}

\subsection{More iterations on 6-d Rosenbrock for GP and PoPBO}
Since 6-d Rosenbrock has a large search space and is hard to converge, we run GP and PoPBO for more iterations (200 queries) and plot the regret in  Fig.~\ref{fig:rosenbrock_many}. We observe that PoPBO consistently outperforms GP after 50 epochs. Moreover, both PoPBO-ERI and PoPBO-R-LCB have lower variance than GP-EI and GP-LCB.

\subsection{Fewer Initial Points on 6-d Rosenbrock}
\begin{figure}
\centering
    \includegraphics[width=0.4\textwidth]{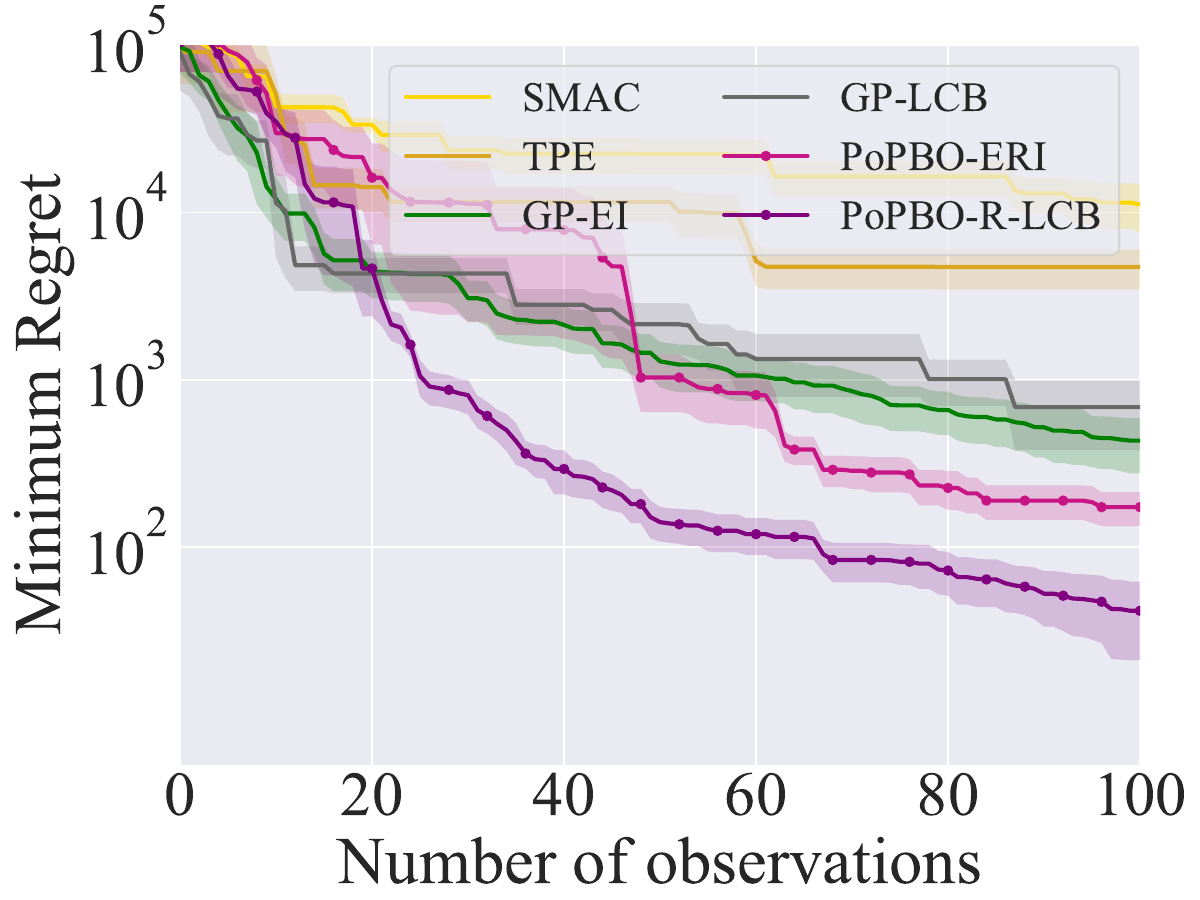} 
\vspace{-12pt}
\caption{Performance on 6-d Rosenbrock with fewer initial points (dimension + 1 = 7 points). For each setting, we conduct replicated experiments for six times with various random seeds.}
\label{fig:rosenbrock_less_init}
\vspace{-10pt}
\end{figure}

We set the number of initial points to 30 for a better preview of the Rosenbrock landscape for all methods. To demonstrate the sample efficiency of our algorithm, we use fewer initial points (7 points for each algorithm) and compare PoPBO with GP-based BO, SMAC and TPE on 6-d Rosenbrock benchmark. The results can be found in Fig.~\ref{fig:rosenbrock_less_init}.

\subsection{Robustness to various noise level}
In our settings, the observations in the simulation function are noiseless, while the observations in the real-world benchmark (HPO-Bench and NAS-Bench-201) are noisy.
For HPO-Bench, the performance of each configuration (hyperparameters of FCNet) is evaluated 4 times under different random seeds. In the experiment settings of HPO-Bench, the average performance is used as the observation, which is naturally noisy.
NAS-Bench-201 attempts to search for a neural architecture that performs best after 200 training epochs. However, following the experiment settings of NAS-Bench-201, only the validation accuracy after 12 training epochs of each architecture can be queried, making the observations noisy.

To verify the robustness to noise of PoPBO on simulated benchmarks, we add Gaussian noises with zero mean and various standard deviation ($\sigma$) to Hartmann, Branin, and Rosenbrock simulation functions and run GP-BO and PoPBO separately. Trends of average regret among six parallel tests are plotted in Fig.~\ref{fig:noise}, showing that GP performs worse with the increment of noise level. In contrast, PoPBO performs much more stable. Moreover, PoPBO outperforms GP when the objective function has large noise ($\sigma=0.1$ for Hartmann, $\sigma=5$ for Branin and Rosenbrock). The results demonstrate the robustness of PoPBO to noise.

\section{Notations}

\begin{itemize}
\item[1] $X$ is the whole feasible domain (search space). If $X$ is continuous, $|X|$ is the volume of $X$. While $X$ is discrete, $|X|$ is the cardinality of it.
\item[2] $S_x = \{y|y\in S, f(y)<f(x)\}$ is the set of better points than $x$ in $S\subset X$, where $S$ can be any continuous domain or discrete set.
\item[3] $\hat{S}$ is a discrete set containing both initial samples for BO and the history queries.
\item[4] Given a specific $S$, $\hat{R}_x(S)$ is a random variable denoting possible ranking of $x$ over a discrete set $\hat{S}_x\cap\hat{S}$. Hence, $\hat{R}_x(S)$ depends on $\hat{S}$, we utilize a hat symbol $\hat{}$ on $R_x$ to omit $\hat{S}$ for conciseness.

\end{itemize}

\section{Broader Impact and Limitations}
This paper addresses the problem of Bayesian Optimization (BO) to enable efficient and effective black-box optimization. It has broad applications in perception tasks, especially computer vision, robotic control and biology. This, on the one hand, would facilitate our daily life, and on the other hand, we shall be careful about their abuse which may break one’s privacy. In this sense, privacy-protection BO is also needed for development, and our techniques can also be of specific help for its generality.

\end{document}